\pdfoutput=1

\documentclass[11pt]{article}

\usepackage[arxiv]{emnlp2021}  

\usepackage{times}
\usepackage{latexsym}

\usepackage[T1]{fontenc}

\usepackage[utf8]{inputenc}

\usepackage{microtype}

\usepackage{amsmath}
\usepackage{tabularx}
\usepackage{booktabs}
\usepackage{siunitx}
\usepackage{multicol}
\usepackage{multirow}
\usepackage{colortbl}
\usepackage{hhline}
\usepackage{xspace}

\usepackage{xcolor}
\definecolor{CustomGray}{gray}{0.95}
\newcolumntype{a}{>{\columncolor{CustomGray}}c}

\newcommand\Tstrut{\rule{0pt}{2.6ex}}         

\usepackage{adjustbox}
\newcolumntype{R}[2]{%
    >{\adjustbox{angle=#1,lap=\width-(#2)}\bgroup}%
    l%
    <{\egroup}%
}

\usepackage{graphicx}
\graphicspath{{images/}}

\newcommand{\fone}{F$_\text{1}$}
\newcommand{\robertal}{RoBERTa$_\text{Large}$}
\newcommand{\rsquad}{R$_\text{SQuAD}$}
\newcommand{\raqa}{R$_\text{SQuAD+AQA}$}
\newcommand{\synqa}{SynQA}
\newcommand{\synqaext}{SynQA$_\text{Ext}$}
\newcommand{\dataset}[1]{\ensuremath{\mathcal{D_{\mathrm{#1}}}}}
\newcommand{\datasetsplit}[2]{\dataset{#1}$^{#2}$}
\newcommand{\squad}{SQuAD}
\newcommand{\squadone}{SQuAD1.1}

\newcommand{\std}[1]{$_{\text{\thinspace{#1}}}$}
\newcommand{\resultsemph}[1]{\underline{#1}}

\definecolor{Maroon}{cmyk}{0, 0.87, 0.68, 0.32}
\definecolor{lightgrey}{rgb}{0.875, 0.875, 0.875}
\definecolor{darkgreen}{rgb}{0.01, 0.75, 0.24}
\definecolor{lightgreen}{rgb}{0.63, 0.85, 0.61}
\definecolor{lightred}{rgb}{0.95, 0.59, 0.61}
\definecolor{csplit}{rgb}{0.45, 0.45, 0.45}

\newcommand{\veryshortarrow}[1][3pt]{\mathrel{%
   \hbox{\rule[\dimexpr\fontdimen22\textfont2-.2pt\relax]{#1}{.4pt}}%
   \mkern-4mu\hbox{\usefont{U}{lasy}{m}{n}\symbol{41}}}}
\def \arrow{$\veryshortarrow$}
\newcommand{\reducedstrut}{\vrule width 0pt height .9\ht\strutbox depth .9\dp\strutbox\relax}
\newcommand{\colbox}[2]{  \begingroup
  \setlength{\fboxsep}{0pt}%
  \colorbox{#1}{\reducedstrut#2\/}%
  \endgroup}
  
\definecolor{caddback}{rgb}{0.90, 0.98, 0.96}
\definecolor{cadd}{rgb}{0, 0.47, 0.34}
\definecolor{cdelback}{rgb}{1, 0.94, 0.92}
\definecolor{cdel}{rgb}{0.83, 0.32, 0.16}
\newcommand{\add}[1]{\colbox{caddback}{\textcolor{cadd}{#1\xspace}}}
\newcommand{\remove}[1]{\colbox{cdelback}{\textcolor{cdel}{#1\xspace}}}
\newcommand{\swap}[2]{\remove{#1} \arrow\add{#2}}

\definecolor{clabelbox}{rgb}{0.25, 0.25, 0.25}

\newcommand{\mybox}[1]{\colbox{lightgrey}{\textcolor{black}{#1}}}
\def \inv{\mybox{INV}}

\def\tabprespace{\vskip -2.4mm}
\def\tabpostspace{\vskip -2.1mm}

\newcommand{\gap}{\vspace{0.2mm}}
\newcommand{\exampleSquad}[4]{
    \gap
    $\begin{array}{l}
    \parbox{12.1cm}{\textbf{C:} #1} \\[2pt]
    \parbox{12.1cm}{
        \textbf{Q:} #2 
        \textbf{A:}\mybox{#3}
        ~~\textbf{M:}  #4
    }
    \end{array}$
    \gap
}

\newcommand{\exampleSquadMax}[4]{
    \parbox{0.88\textwidth}{
        \vspace{3pt}
        \textbf{C:} #1 \\
        \textbf{Q:} #2 
        \textbf{A:}\add{#3} ~~\textbf{M:}\remove{#4}
        \vspace{3pt}
    }
}

\newcommand*{\affmark}[1][*]{\textsuperscript{#1}}

\title{Improving Question Answering Model Robustness with \\Synthetic Adversarial Data Generation}

\author{
  \textbf{Max Bartolo}\affmark[$\dagger$]\thanks{\ \ Most of this work was carried out while MB was an intern at at Facebook~AI~Research.} \quad \textbf{Tristan Thrush}\affmark[$\ddagger$] \quad \textbf{Robin Jia}\affmark[$\ddagger$] \quad \textbf{Sebastian Riedel}\affmark[$\dagger\ddagger$] \\
  \textbf{Pontus Stenetorp}\affmark[$\dagger$] \quad \textbf{Douwe Kiela}\affmark[$\ddagger$] \\
  $^{\dagger}$University College London \quad $^{\ddagger}$Facebook AI Research \\
  \texttt{m.bartolo@cs.ucl.ac.uk} \\
}

\begin{document}
\maketitle

\begin{abstract}


Despite recent progress, state-of-the-art question answering models remain vulnerable to a variety of adversarial attacks. 
While dynamic adversarial data collection, in which a human annotator tries to write examples that fool a model-in-the-loop, can improve model robustness, this process is expensive which limits the scale of the collected data.
In this work, we are the first to use synthetic adversarial data generation to make question answering models more robust to human adversaries.
We develop a data generation pipeline that selects source passages, identifies candidate answers, generates questions, then finally filters or re-labels them to improve quality.
Using this approach, we amplify a smaller human-written adversarial dataset to a much larger set of \textit{synthetic} question-answer pairs.
By incorporating our synthetic data, we improve the state-of-the-art on the AdversarialQA dataset by 3.7\fone{} and improve model generalisation on nine of the twelve MRQA datasets.
We further conduct a novel human-in-the-loop evaluation and show that our models are considerably more robust to new human-written adversarial examples:
crowdworkers can fool our model only $8.8\%$ of the time on average, compared to $17.6\%$ for a model trained without synthetic data.
\end{abstract}

\section{Introduction}

\begin{figure}[t]
    \centering
    \includegraphics[width=\columnwidth]{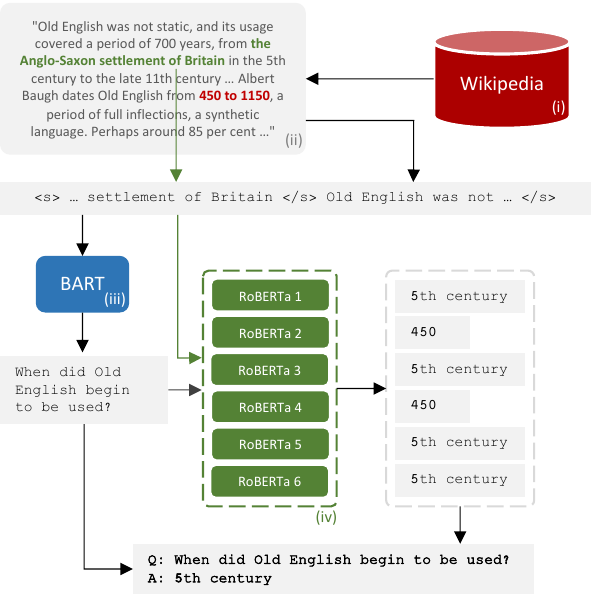}
    \caption{The Synthetic Adversarial Data Generation Pipeline showing: (i) passage selection from Wikipedia; (ii) answer candidate selection and filtering by model confidence (an example retained answer shown in green, and a dropped answer candidate in red); (iii) question generation using BART$_\text{Large}$; and (iv) answer re-labelling using self-training. The generated synthetic data is then used as part of the training data for a downstream Reading Comprehension model.
    } 
    \label{fig:splash}
\end{figure}

Large-scale labelled datasets like SQuAD~\cite{rajpurkar2016squad} and SNLI~\cite{bowman-etal-2015-large} have been driving forces in natural language processing research.
Over the past few years, however, such ``statically collected’’ datasets have been shown to suffer from various problems. In particular, they often exhibit inadvertent spurious statistical patterns that models learn to exploit, leading to poor model robustness and generalisation~\cite{jia2017adversarial,gururangan2018annotation,geva-etal-2019-modeling,mccoy-etal-2019-right,lewis2020question}.
A recently proposed alternative is dynamic data collection \cite{bartolo2020beat,nie-etal-2020-adversarial}, where data is collected with both humans and models in the annotation loop. Usually, these humans are instructed to ask adversarial questions that fool existing models.
Dynamic adversarial data collection is often used to evaluate the capabilities of current state-of-the-art models, but it can also create higher-quality training data~\cite{bartolo2020beat,nie-etal-2020-adversarial} due to the added incentive for crowdworkers to provide challenging examples. 
It can also reduce the prevalence of dataset biases and annotator artefacts over time~\cite{bartolo2020beat,nie-etal-2020-adversarial}, since such phenomena can be subverted by model-fooling examples collected in subsequent rounds.
However, dynamic data collection can be more expensive than its static predecessor as creating examples that elicit a certain model response (i.e., fooling the model) requires more annotator effort, resulting in more time spent, and therefore higher cost per example.
In this work, we develop a synthetic adversarial data generation pipeline, making novel contributions to the answer selection, question generation, and filtering and re-labelling tasks.
We show that dynamic adversarial data collection can be made more sample efficient by synthetically generating (see Figure~\ref{fig:splash}) examples that improve the robustness of models
in terms of performance on adversarially-collected datasets, comprehension skills, and domain generalisation. 
We are also the first to evaluate models in-the-loop for robustness to human adversaries using the \textit{macro-averaged validated model error rate}, demonstrating considerable improvements with crowdworkers only able to fool the model-in-the-loop 8.8\% of the time on average, compared to 17.6\% for our best baseline.
The collected dataset will form part of the evaluation for a new round of the Dynabench QA task.\footnote{\url{https://dynabench.org/tasks/qa}}
%

\section{Related Work}

\subsection{Adversarial Data Collection}
We directly extend the AdversarialQA dataset collected in ``Beat the AI''~\citep{bartolo2020beat}, which uses the same passages as \squadone{}.
AdversarialQA was collected by asking crowdworkers to write extractive question-answering examples that three different models-in-the-loop were unable to answer correctly, creating the \dataset{BiDAF}, \dataset{BERT}, and \dataset{RoBERTa} subsets. 
Other datasets for question answering~\citep{rajpurkar-etal-2018-know,dua-etal-2019-drop,wallace-etal-2019-trick}, sentiment analysis~\citep{potts-etal-2020-dynasent}, hate speech detection \citep{vidgen2020learning}, and natural language inference \citep{nie-etal-2020-adversarial} have been collected in a similar manner.
While appealing, human-generated adversarial data is expensive to collect;
our work is complementary in that it explores methods to extract further value from existing adversarially collected datasets without requiring additional annotation effort.
\subsection{Synthetic Question Generation}
Many approaches have been proposed to generate question-answer pairs given a passage~\citep{du-etal-2017-learning,du-cardie-2018-harvesting,zhao-etal-2018-paragraph,lewis2018generative,alberti-etal-2019-synthetic,puri-etal-2020-training,lewis2021paq}.
These generally use a two-stage pipeline that first identifies an answer conditioned on a passage, then generates a question conditioned on the passage and answer; we train a similar pipeline in our work. 
\mbox{G-DAUG}~\citep{yang-etal-2020-generative}
trains generative models to synthesise training data for commonsense reasoning.
Our work focuses on extractive question-answering~(QA), which motivates the need for different generative models.
\citet{yang-etal-2020-generative} filter generated examples using influence functions, or methods that attempt to maximise diversity;
we find that a different approach that considers answer agreement between QA models trained with different random seeds leads to better performance in our setting. 

\subsection{Self-training}
In self-training, a model is trained to both predict correctly on labelled examples and increase its confidence on unlabelled examples.
Self-training can yield complementary accuracy gains with pretraining \citep{du2020selftraining}
and can improve robustness to domain shift \citep{kumar2020gradual}.
In our setting, large amounts of unlabelled adversarial-style questions are not readily available, which motivates our use of a question generation model.

\subsection{Human Evaluation}
The ultimate goal of automatic machine learning model evaluation is usually stated as capturing human judgements \cite{callison2006re,hill2015simlex,vedantam2015cider,liu2016not}.
Evaluation with real humans is considered beneficial, but not easily scalable, and as such is rarely conducted in-the-loop.
With NLP model capabilities ever improving, adversarial \emph{worst case} evaluation becomes even more pertinent.
To our knowledge, this work is the first to compare models explicitly by their adversarial validated model error rate (vMER), which we define in Section~\ref{sec:human_eval}.



\section{Synthetic Data Generation}
We develop a synthetic data generation pipeline for QA that involves four stages: passage selection, answer candidate selection, question generation, and synthetic data filtering and re-labelling.
Due to the complexity of the system, we study each of these in isolation, and then combine our best identified approaches for the final systems.
We evaluate each component both intrinsically and on their contribution to downstream QA performance on the AdversarialQA test sets and an unseen split of the \squadone{} dev set.
The final synthetic data generation pipeline consists of:
\begin{enumerate}
    \item \textit{Passage selection}:  we use passages from Wikipedia for this work.
    \item \textit{Answer Candidate selection}: the model identifies spans within the passage that are likely to be answers to a question.
    \item \textit{Question Generation}: a generative model is used to generate a question, conditioned on the passage and each answer.
    \item \textit{Filtering and Re-labelling}: synthetic question-answer pairs that do not meet the necessary criteria are discarded, or have their answers re-labelled using self-training.
\end{enumerate}
Results for the baseline and overall best performing systems are shown in Table~\ref{tab:final_model_results}. 
Results for ELECTRA\textsubscript{Large}~\citep{Clark2020ELECTRA} showing further performance gains are in Appendix~\ref{appendix:results_electra}.
%

\subsection{Data Generation Pipeline}

%
\begin{table}[t]
    \centering
    \footnotesize
    \setlength{\tabcolsep}{0pt}
        \begin{tabular} {@{\extracolsep{0pt}}l r@{\extracolsep{8pt}}r@{\extracolsep{8pt}}r@{}}
                \textbf{Model} & \textbf{Precision (\%)} & \textbf{Recall (\%)} &\textbf{\fone{} (\%)} \\
        \toprule
        
        POS Extended & 12.7 & 65.2 & 20.7 \\
        Noun Chunks & 17.4 & 36.9 & 22.5 \\
        Named Entities & 30.3 & 30.0 & 27.1 \\
        Span Extraction, \textit{k}=15 & 22.5 & 26.6 & 23.7 \\
        BART$_\text{ans. only}$, \textit{k}=15 & 27.7 & 31.3 & 28.6 \\
        SAL (ours) & 28.6 & 44.2 & \textbf{33.7} \\
        
        \bottomrule
        \end{tabular}
    \caption{
    Answer selection results on aligned test set.
    }
    \label{tab:results_ans_select}
\end{table}
%

%
\begin{table*}[t]
    \aboverulesep=0pt
    \belowrulesep=0pt
    \renewcommand{\arraystretch}{1.2}
    \footnotesize
    \centering
    \setlength{\tabcolsep}{11.2pt}
        \begin{tabular} {@{\extracolsep{1pt}}l | c | ac ac ac ac }
                \multirow{2}{0pt}{\textbf{Method}} & 
                \multirow{2}{2.2em}{\textbf{\#QA pairs}} & 
                \multicolumn{2}{c}{\dataset{SQuAD}} & \multicolumn{2}{c}{\textbf{\dataset{BiDAF}}} & \multicolumn{2}{c}{\textbf{\dataset{BERT}}} &\multicolumn{2}{c}{\textbf{\dataset{RoBERTa}}} \\
                \hhline{~|~|--||--||--||--}
             &  &\emph{EM}&\emph{F$_\text{1}$}&\emph{EM}&\emph{F$_\text{1}$}&\emph{EM}&\emph{F$_\text{1}$} &\emph{EM}&\emph{F$_\text{1}$} \\
             
        \toprule
        
        POS Extended & 999,034 & 53.8 & 71.4 & 32.7 & 46.9 & 30.8 & 40.2 & 20.4 & 27.9 \\
        Noun Chunks & 581,512 & 43.3 & 63.7 & 28.7 & 43.1 & 22.3 & 31.4 & 18.2 & 27.4 \\
        Named Entities & 257,857 & 54.2 & 69.7 & 30.5 & 42.5 & 26.6 & 35.4 & 18.1 & 24.0 \\
        Span Extraction & 377,774 & 64.7 & 80.1 & 37.8 & 53.9 & 27.7 & 39.1 & 16.7 & 26.9 \\
        SAL (ours) & 566,730 & 68.2 & \textbf{82.6} & 43.2 & 59.3 & 34.9 & 45.4 & \textbf{25.2} & \textbf{32.8} \\
        SAL threshold (ours) & 393,164 & \textbf{68.5} & 82.0 & \textbf{46.0} & \textbf{60.3} & \textbf{36.5} & \textbf{46.8} & 24.2 & 32.4 \\

        \bottomrule
        \end{tabular}
    \caption{Downstream test results for a RoBERTa$_\text{Large}$ QA model trained on synthetic data generated using different answer selection methods combined with a BART$_\text{Large}$ 
    question generator (trained on SQuAD$_{10k}$ + \dataset{AQA}). 
    }
    \label{tab:results_ans_select_downstream}
\end{table*}

%

%
In order to generate synthetic adversarial examples, we first select passages, then identify candidate answers in those passages, generate corresponding questions for these answers, and then filter or re-label for improved quality based on various criteria.
%

\subsubsection{Passage Selection}
The text passages we use are sourced from SQuAD (further details can be found in Appendix~\ref{appendix:passage_selection}).
We also experiment with using passages external to SQuAD, which are also sourced from Wikipedia.
To preserve evaluation integrity, we analyse the 8-gram overlap of all external passages to the evaluation datasets, after normalisation to lower-cased alphanumeric words with a single space delimiter~\cite{radford2019language}.
%
We find that just 0.3\% of the external passages have any overlap with the evaluation sets, and filter these out.
%

\subsubsection{Answer Candidate Selection}
\label{sec:answer_candidate_selection}

The next step is to identify which spans of text within the passages are likely to be answers to a question.
We investigate a range of existing methods for answer candidate selection, which takes the passage as input and outputs a set of possible answers. We further propose a self-attention-based classification head that jointly models span starts and ends, with improved performance.
Since SQuAD and the AdversarialQA datasets use the same passages partitioned into the same data splits, we align the annotated answers to create representative answer selection training, validation and test sets. 
Dataset statistics (see Appendix~\ref{appendix:ans_candidate}), highlight the high percentage of overlapping answers suggesting that existing answer tagging methods~\cite{Zhou2017NeuralQG, zhao-etal-2018-paragraph} might struggle, and models should ideally be capable of handling span overlap.
\paragraph{Baseline Systems}{
    We investigate three baseline systems; noun phrases and named entities following~\citet{lewis-etal-2019-unsupervised}, as well as an extended part-of-speech tagger incorporating named entities, adjectives, noun phrases, numbers, distinct proper nouns, and clauses.
}

\paragraph{Span Extraction}{
    We fine-tune a \robertal{} span extraction model as investigated in previous work~\cite{alberti-etal-2019-synthetic, lewis2018generative}.
    We treat the number of candidates to sample as a hyper-parameter and select the optimal value for $k \in \{1, 5, 10, 15, 20\}$ on the validation set.
}

\paragraph{Generative Answer Detection}{
    We use BART$_\text{Large}$~\cite{lewis-etal-2020-bart} in two settings; one generating answer and question, and the other where we generate the answer only, as we find that this setting provides better control of answer diversity.
    We use the same range of $k \in \{1, 5, 10, 15, 20\}$ for both settings.
}

\paragraph{Self-Attention Labelling (SAL)}{
    We propose a multi-label classification head to jointly model candidate start and end tokens, and provide a binary label for whether each possible span of text from the passage is a candidate answer.
    We adapt scaled dot-product attention~\cite{vaswani2017attention} where the candidate start, $\textbf{S}$, and end, $\textbf{E}$, token representations are analogous to the projected layer input queries and keys.
    We apply a sigmoid over the computed attention scores, giving a matrix where each cell gives the probability $p(a_{ij} | c)$ of whether the span in the context, $c$, with start index $i$ and end index $j$ is a valid answer candidate. Formally:
    $$
    p(a_{ij} | c) = \sigma \left( \frac{\sum_{k=1}^d s_{ik}e_{kj}}{\sqrt{d}} \right)
    $$
    We optimise using binary cross-entropy, masking out impossible answer spans defined as those not in the passage, with end indices before start, or longer than the maximum permitted answer length, and upweigh positive examples to help counteract the class imbalance.
    We decode from the output probability matrix to the original passage tokens using a reversible tokeniser and use a probability threshold of $0.5$ for candidate selection, which can be adapted to tune precision and recall.
    While answer candidate selection only requires a single attention head, the multi-head implementation allows application to any labelling task requiring span modelling with overlaps, where each head is trained to predict labels for each class, such as for nested Named Entity Recognition.
    We implement this in \textit{Transformers}~\cite{wolf-etal-2020-transformers} and fine-tune \robertal{} with SAL on the answer selection dataset.
}

\paragraph{Evaluation} {
    We evaluate performance on the answer selection dataset using entity-level precision, recall, and \fone{} on unique normalised candidates.
    Results are shown in Table~\ref{tab:results_ans_select}.
    We further investigate the effects of different answer candidate selection methods on downstream QA model performance (see Table~\ref{tab:results_ans_select_downstream}) by training a RoBERTa$_\text{Large}$ model on synthetic QA pairs generated when using different answer selection methods.
    To eliminate generated dataset size as a potential confounder, we also replicate these experiments using a sample of 87,000 examples and find similar results (see Appendix~\ref{appendix:ans_candidate}).
}

\subsubsection{Question Generation}
Once answer candidates have been identified for a selected passage, we then generate a corresponding question by directly fine-tuning a BART$_{\text{Large}}$~\citep{lewis-etal-2020-bart} autoregressive sequence generation decoder.\footnote{We also try generating multiple questions but consistently find that generating one question per answer provides the best downstream results despite the additional data.}
To discourage the model from memorising the questions in the SQuAD training set and directly reproducing these, we train on a subset of 10k examples from SQuAD, selected such that they correspond to the same source passages as the AdversarialQA training data.
This ensures that when scaling up synthetic generation, the vast majority of passages are previously completely unseen to the generator.
%

%
\begin{table}[t]
    \renewcommand{\arraystretch}{1.2}
    \footnotesize
    \setlength{\tabcolsep}{3pt}
        \begin{tabular} {@{\extracolsep{0pt}} m{0.18\columnwidth} | m{0.79\columnwidth} @{}}
        
        \multicolumn{2}{c}{\parbox{0.97\columnwidth}{\textbf{Context:} \textit{Following the series revival in 2005, \remove{Derek Jacobi~\textsubscript{\textbf{ANS}}} provided the character's re-introduction in the 2007 episode "Utopia". During that story the role was then assumed by John Simm who returned to the role multiple times through the Tenth Doctor's tenure. As of the 2014 episode "Dark Water," it was revealed that the Master had become a female incarnation or "Time Lady," going by the name of "Missy", played by Michelle Gomez.}}} \\

        \midrule
        \squad{}$_{10k}$ & Who portrayed the Master in the 2007 episode "Utopia"? \\
        \midrule
        \dataset{BiDAF} & Who replaced John Simm as the Tenth Doctor?	\textbf{(Answer Mismatch)} \\
        \midrule
        \dataset{BERT} & Who played the Master in the 2007 episode "Utopia"?	 \\
        \midrule
        \dataset{RoBERTa} & Who was the first actor to play the Master? \\
        \midrule
        \dataset{AQA} & Who played the Master first, Derek Jacobi or John Simm? \\
        \midrule
        \squad{}$_{10k}$ + \dataset{AQA} & Who re-introduced the character of the Master? \\
        \bottomrule
        \end{tabular}
    \caption{Examples of questions generated using BART trained on different source datasets.}
    \label{tab:examples_qgen}
\end{table}
\begin{table}[t]
    \centering
    \footnotesize
    \setlength{\tabcolsep}{2pt}
        \begin{tabular} {@{\extracolsep{0pt}} 
        p{0.28\columnwidth}
        >{\raggedleft\arraybackslash}p{0.15\columnwidth}
        >{\raggedleft\arraybackslash}p{0.18\columnwidth} >{\raggedleft\arraybackslash}p{0.18\columnwidth}
        >{\raggedleft\arraybackslash}p{0.15\columnwidth} @{}}
                \textbf{Model} & \textbf{Valid} & \textbf{Answer Mismatch} & \textbf{Ungramm-atical} & \textbf{Invalid} \\
        \toprule
        
        \squad{}$_{10k}$ & 90.0\% & 10.0\% & 0.0\% & 0.0\% \\
        \dataset{BiDAF} & 70.0\% & 30.0\% & 0.0\% & 0.0\% \\
        \dataset{BERT} & 76.7\% & 23.3\% & 0.0\% & 0.0\% \\
        \dataset{RoBERTa} & 70.0\% & 20.0\% & 0.0\% & 10.0\% \\
        \dataset{AQA} & 76.7\% & 16.7\% & 0.0\% & 6.7\% \\
        \squad{}$_{10k}$+\dataset{AQA} & 93.3\% & 6.7\% & 0.0\% & 0.0\% \\

        \bottomrule
        \end{tabular}
    \caption{
    Manual analysis of questions generated when training on different source data.
    }
    \label{tab:results_qgen_manual}
\end{table}
%

\paragraph{Source Questions}{
Since the types of questions a generative model is trained on can impact both performance and diversity, we experiment with training on SQuAD and different subsets of AdversarialQA, and the combination of both. 
Examples of the generated questions are shown in Table~\ref{tab:examples_qgen}.
We carry out a manual answerability analysis on a random sample of 30 generated questions (using beam search with $k=5$) in each of these settings (see Table~\ref{tab:results_qgen_manual} and Appendix~{\ref{appendix:answerability_analysis}}).
We define answerability by the following criteria: (i) The question must be answerable from a single continuous span in the passage; (ii) There must be only one valid (or clearly one most valid) answer (e.g. in the case of a co-reference the canonical entity name should be the answer); (iii) A human should be able to answer the question correctly given sufficient time; and (iv) The correct answer is the one on which the model was conditioned during question generation.
We find that when the models attempt to generate complex questions, the generated question is often inconsistent with the target answer, despite remaining well-formed.
We also observe that when the generated question requires external knowledge (e.g. ``What is a tribe?'' or ``Which is not a country?'') the models are reasonably consistent with the answer, however, they often lose answer consistency when answering the question requires resolving information in the passage (e.g. ``What is the first place mentioned?'').
}

%
\begin{table*}[t]
    \aboverulesep=0pt
    \belowrulesep=0pt
    \renewcommand{\arraystretch}{1.2}
    \footnotesize
    \centering
    \setlength{\tabcolsep}{11.3pt}
        \begin{tabular} {@{\extracolsep{1pt}}l | r | ac ac ac ac }
                \multirow{2}{0pt}{\textbf{Method}} & 
                \multirow{2}{2.2em}{\textbf{\#QA pairs}} &  \multicolumn{2}{c}{\dataset{SQuAD}} & \multicolumn{2}{c}{\textbf{\dataset{BiDAF}}} & \multicolumn{2}{c}{\textbf{\dataset{BERT}}} &\multicolumn{2}{c}{\textbf{\dataset{RoBERTa}}} \\
                \hhline{~|~|--||--||--||--}
             &  &\emph{EM}&\emph{F$_\text{1}$}&\emph{EM}&\emph{F$_\text{1}$}&\emph{EM}&\emph{F$_\text{1}$} &\emph{EM}&\emph{F$_\text{1}$} \\
             
        \toprule
        
        \rsquad{} & 87,599 & 73.2 & 86.3 & 48.9 & 64.3 & 31.3 & 43.5 & 16.1 & 26.7 \\
        
        \raqa{} & 117,599 & \resultsemph{74.2} & \resultsemph{86.9} & \resultsemph{57.4} & \resultsemph{72.2} & \resultsemph{53.9} & \resultsemph{65.3} & \resultsemph{43.4} & \resultsemph{54.2} \\
        
        \midrule
        
        \squad{}$_{10k}$ & 87,598 & 69.2 & 82.6 & 37.1 & 52.1 & 22.4 & 32.3 & 13.9 & 22.3 \\
        
        \dataset{BiDAF} & 87,598 & 67.1 & 80.4 & 41.4 & 56.5 & \textbf{33.1} & 43.8 & 22.0 & 32.5 \\
        
        \dataset{BERT} & 87,598 & 67.4 & 80.2 & 36.3 & 51.1 & 30.3 & 40.6 & 18.8 & 29.5 \\
        
        \dataset{RoBERTa} & 87,598 & 63.4 & 77.9 & 32.6 & 47.9 & 27.2 & 37.5 & 20.6 & 32.0 \\
        
        \dataset{AQA} & 87,598 & 65.5 & 80.1 & 37.0 & 53.0 & 31.1 & 40.9 & \textbf{23.2} & \textbf{33.3} \\
        
        \squad{}$_{10k}$ + \dataset{AQA} & 87,598 & \textbf{71.9} & \textbf{84.7} & \textbf{44.1} & \textbf{58.8} & 32.9 & \textbf{44.1} & 19.1 & 28.8 \\

        \bottomrule
        \end{tabular}
    \caption{Downstream QA test results using generative models trained on different source data. We compare these results to baseline RoBERTa models trained on SQuAD, and on the combination of SQuAD and AdversarialQA.}
    \label{tab:results_qgen_downstream}
\end{table*}
%

%
For each of these models, we generate 87k examples (the same size as the SQuAD training set to facilitate comparison) 
using the human-provided answers,
and then measure the effects on downstream performance by training a QA model on this synthetic data.
Results are shown in Table~\ref{tab:results_qgen_downstream}.
We find that, in this setting, the best source data for the generative model is consistently the combination of SQuAD and AdversarialQA.
We also note that using only synthetic generated data, we can achieve good performance on \dataset{SQuAD} consistent with the findings of \citet{puri-etal-2020-training}, and outperform the model trained on the human-written SQuAD data on \dataset{BERT} (+0.6\fone{}) and \dataset{RoBERTa} (+6.6\fone{}).
This is in line with the observations of \citet{bartolo2020beat} suggesting that the distribution of the questions collected using progressively stronger models-in-the-loop is less similar to that of SQuAD.
It also shows that the generator can successfully identify and reproduce patterns of adversarially-written questions.
However, the results using synthetic data alone are considerably worse than when training the QA model on human-written adversarial data with, for example, a performance drop of 21.2\fone{} for \dataset{BERT}.
This suggests that while we can do well on SQuAD using synthetic questions alone, we may need to combine the synthetic data with the human-written data for best performance in the more challenging adversarial settings.

\paragraph{Question Diversity}{
In order to provide training signal diversity to the downstream QA model, we experiment with a range of decoding techniques (see Appendix~\ref{appendix:further_question_diversity}), and then evaluate these by downstream performance of a QA model trained on the questions generated in each setting.
We observe minimal variation in downstream performance as a result of question decoding strategy, with the best downstream results obtained using nucleus sampling ($top_p = 0.75$). However, we also obtain similar downstream results with standard beam search using a beam size of 5.
We find that, given the same computational resources, standard beam search is roughly twice as efficient, and therefore opt for this approach for our following experiments.

}

\subsubsection{Filtering and Re-labelling}

%
\begin{table*}[t]
    \aboverulesep=0pt
    \belowrulesep=0pt
    \renewcommand{\arraystretch}{1.2}
    \centering
    \footnotesize
    \setlength{\tabcolsep}{6.0pt}
        \begin{tabular} {@{\extracolsep{1pt}} p{0.38\textwidth} | c | ac ac ac ac }
                \multirow{2}{10em}{\textbf{Filtering Method}} & 
                \multirow{2}{2.2em}{\textbf{\#QA pairs}} &  \multicolumn{2}{c}{\dataset{SQuAD}} & \multicolumn{2}{c}{\textbf{\dataset{BiDAF}}} & \multicolumn{2}{c}{\textbf{\dataset{BERT}}} &\multicolumn{2}{c}{\textbf{\dataset{RoBERTa}}} \\
                \hhline{~|~|--||--||--||--}
             &  &\emph{EM}&\emph{F$_\text{1}$}&\emph{EM}&\emph{F$_\text{1}$}&\emph{EM}&\emph{F$_\text{1}$} &\emph{EM}&\emph{F$_\text{1}$} \\
             
        \toprule
        
        Answer Candidate Conf. ($thresh = 0.6$) & 362,281 & 68.4 & 82.4 & 42.9 & 57.9 & 36.3 & 45.9 & 28.0 & 36.5 \\
        
        Question Generator Conf. ($thresh = 0.3$) & 566,725 & 69.3 & 83.1 & 43.5 & 58.9 & 36.3 & 46.6 & 26.2 & 34.8 \\
        
        Influence Functions & 288,636 & 68.1 & 81.9 & 43.7 & 58.6 & 36.1 & 46.6 & 27.4 & 36.4 \\
        
        Ensemble Roundtrip Consistency ($6/6$ correct) & 250,188 & 74.2 & 86.2 & 55.1 & 67.7 & 45.8 & 54.6 & 31.9 & 40.3 \\
        
        Self-training (ST) & 528,694 & 74.8 & 87.0 & 53.9 & 67.9 & 47.5 & 57.6 & 35.2 & 44.6 \\
        
        Answer Candidate Conf. ($thresh = 0.5$) \& ST & 380,785 & \textbf{75.1} & \textbf{87.0} & \textbf{56.5} & \textbf{70.0} & \textbf{47.9} & \textbf{58.7} & \textbf{36.0} & \textbf{45.9} \\

        \bottomrule
        \end{tabular}
    \caption{Downstream QA test results for different filtering strategies, showing best hyper-parameter settings.}
    \label{tab:results_qgen_filtering}
\end{table*}
%

%
The synthetic question generation process can introduce various sources of noise, as seen in the previous analysis, which could negatively impact downstream results.
To mitigate these effects, we explore a range of filtering and re-labelling methods. Results for the best performing hyper-parameters of each method are shown in Table~\ref{tab:results_qgen_filtering} and results controlling for dataset size are in Appendix~\ref{appendix:data_size_control}.

\paragraph{Answer Candidate Confidence} {
    We select candidate answers using SAL (see section~\ref{sec:answer_candidate_selection}), and filter based on the span extraction confidence of the answer candidate selection model.
}

\paragraph{Question Generator Confidence} {
    We filter out samples below various thresholds of the probability score assigned to the generated question by the question generation model.
}

\paragraph{Influence Functions} {
    We use influence functions~\citep{cook1982residuals, DBLP:conf/icml/KohL17} to estimate the effect on the validation loss of including a synthetic example as explored by~\citet{yang-etal-2020-generative}, but adapted for QA.
    We filter out examples estimated to increase the validation loss.
}

\paragraph{Ensemble Roundtrip Consistency} {
    Roundtrip consistency~\citep{alberti-etal-2019-synthetic, fang2020accelerating} uses an existing fine-tuned QA model to attempt to answer the generated questions, ensuring that the predicted answer is consistent with the target answer prompted to the generator.
    Since our setup is designed to generate questions which are intentionally challenging for the QA model to answer, we attempt to exploit the observed variation in model behaviour over multiple random seeds, and replace the single QA model with a six-model ensemble.
    We find that filtering based on the number of downstream models that correctly predict the original target answer for the generated question produces substantially better results than relying on the model confidence scores, which could be prone to calibration imbalances across models.
}

\paragraph{Self-training} {
    Filtering out examples that are not roundtrip-consistent can help eliminate noisy data, however, it also results in (potentially difficult to answer) questions to which a valid answer may still exist being unnecessarily discarded.
    Self-training has been shown to improve robustness to domain shift~\citep{kumar2020gradual} and, in our case, we re-label answers to the generated questions based on the six QA model predictions.
    Specifically, in our best-performing setting, we keep any examples where at least five of the six QA models agree with the target answer (i.e. the one with which the question generator was originally prompted), re-label the answers for any examples where at least two of the models QA agree among themselves, and discard the remaining examples (i.e. those for which there is no agreement between any of the QA models).
}

We find that the best method combines self-training with answer candidate confidence filtering.
By using appropriate filtering of the synthetic generated data, combined with the ability to scale to many more generated examples, we approach the performance of \raqa{}, practically matching performance on SQuAD and reducing the performance disparity to just 2.2\fone{} on \dataset{BiDAF}, 6.6\fone{} on \dataset{BERT}, and 8.3\fone{} on \dataset{RoBERTa}, while still training solely on synthetic data.
%

%
\begin{table*}[t]
    \aboverulesep=0pt
    \belowrulesep=0pt
    \renewcommand{\arraystretch}{1.2}
    \centering
    \footnotesize
    \setlength{\tabcolsep}{7.0pt}
        \begin{tabular} {@{\extracolsep{0pt}}p{0.12\textwidth} | p{0.20\textwidth} | ac ac ac | c @{}}
                \multirow{2}{0em}{\textbf{Model}} & 
                \multirow{2}{6em}{\textbf{Training Data}} & \multicolumn{2}{c}{\textbf{\dataset{BiDAF}}} & \multicolumn{2}{c}{\textbf{\dataset{BERT}}} & \multicolumn{2}{c|}{\textbf{\dataset{RoBERTa}}} & \multicolumn{1}{c}{\textbf{mvMER$^*$}} \\
                \hhline{~|~|--|--|--|-}
             &&  \emph{EM}&\emph{\fone{}}&\emph{EM}&\emph{\fone{}}&\emph{EM}&\emph{\fone{}} &\emph{\%} \\
        
        \toprule
            \rsquad{} & SQuAD & 
            48.6\std{1.3} &	64.2\std{1.5} & 30.9\std{1.3} &	43.3\std{1.7} &	15.8\std{0.9} & 26.4\std{1.3} &	20.7\% \\
            
            \raqa{} & \hspace{4pt} $\uparrow$ + AQA & 
            59.6\std{0.5} &	73.9\std{0.5} &	54.8\std{0.7} &	64.8\std{0.9} &	41.7\std{0.6} &	53.1\std{0.8} &	17.6\%  \\
        
        \midrule
        
            \synqa{} & \hspace{14pt} $\uparrow$ + SynQA$_\text{SQuAD}$ & 
            62.5\std{0.9} &	76.0\std{1.0} & 58.7\std{1.4} & 68.3\std{1.4} & 46.7\std{1.8} & \textbf{58.0} \std{1.8} &	\textbf{\hspace{5pt}8.8\%}  \\
            
            \synqaext{} & \hspace{24pt} $\uparrow$ + SynQA$_\text{Ext}$ & 
            \textbf{62.7}\std{0.6} & \textbf{76.2}\std{0.5} & \textbf{59.0}\std{0.7} & \textbf{68.9}\std{0.5} & \textbf{46.8}\std{0.5} & 57.8\std{0.8} & 12.3\%  \\

        \bottomrule
        \end{tabular}
    \caption{Test set results for \robertal{} trained on different datasets, and augmented with synthetic data. AQA is the AdversarialQA data consisting of the combined \dataset{BiDAF}, \dataset{BERT}, and \dataset{RoBERTa} from~\citet{bartolo2020beat}. We report the mean and standard deviation (subscript) over 6 runs with different random seeds. mvMER is the macro-averaged validated model error rate in the adversarial human evaluation setting ($^*$lower is better).}
    \label{tab:final_model_results}
\end{table*}
%

\subsection{End-to-end Synthetic Data Generation}
We also try using BART to both select answers and generate questions in an end-to-end setting. We experiment with different source datasets, number of generations per passage, and decoding hyper-parameters, but our best results fall short of the best pipeline approach at 62.7/77.9 EM/\fone{} on \dataset{SQuAD}, 30.8/47.4 on \dataset{BiDAF}, 23.6/35.6 on \dataset{BERT}, and 18.0/28.3 on \dataset{RoBERTa}.
These results are competitive when compared to some of the other answer candidate selection methods we explored, however, fall short of the results obtained when using SAL.
We find that this approach tends to produce synthetic examples with similar answers, but leave exploring decoding diversity to future work.
%

\subsection{Fine-tuning Setup}
We investigate two primary fine-tuning approaches: combining all training data, and a two-stage set-up in which we first fine-tune on the generated synthetic data, and then perform a second-stage of fine-tuning on the SQuAD and AdversarialQA human-written datasets.
Similar to~\citet{yang-etal-2020-generative}, we find that two-stage training marginally improves performance over standard mixed training, and we use this approach for all subsequent experiments.
%

\section{Measuring Model Robustness}
Based on the findings in the previous section, we select four final models for robustness evaluation: 
\begin{enumerate}
    \item \rsquad{}: using the SQuAD1.1 training data.
    \item \raqa{}: trained on SQuAD combined and shuffled with AdversarialQA.
    \item \synqa{}: uses a two-stage fine-tuning approach, first trained on 314,811 synthetically generated questions on the passages in the SQuAD training set, and then further fine-tuned on SQuAD and AdversarialQA.
    \item \synqaext{}: first trained on the same synthetic SQuAD examples as (iii) combined with 1.5M synthetic questions generated on the previously described Wikipedia passages external to SQuAD, and then further fine-tuned on SQuAD and AdversarialQA.
\end{enumerate}

Individual models are selected for the best combined and equally-weighted performance on a split of the SQuAD validation set and all three AdversarialQA validation sets.
We first evaluate model robustness using three existing paradigms: adversarially-collected datasets, checklists, and domain generalisation. 
We also introduce adversarial human evaluation, a new way of measuring robustness with direct interaction between the human and model.

\subsection{Adversarially-collected Data}
We evaluate the final models on AdversarialQA, with results shown in Table~\ref{tab:final_model_results}.
We find that synthetic data augmentation yields state-of-the-art results on AdversarialQA, providing performance gains of 2.3\fone{} on \dataset{BiDAF}, 4.1\fone{} on \dataset{BERT}, and 4.9\fone{} on \dataset{RoBERTa} over the baselines while retaining good performance on \squad{}, a considerable improvement at no additional annotation cost. 

\subsection{Comprehension Skills}
CheckList~\citep{ribeiro-etal-2020-beyond} is a model agnostic approach that serves as a convenient test-bed for evaluating what \textit{comprehension skills} a QA model could learn.
We find that some skills that models struggle to learn when trained on \squad{}, such as discerning between profession and nationality, or 
handling negation in questions, can be learnt by incorporating adversarially-collected data during training (see Appendix~\ref{appendix:results_checklist}).
Furthermore, augmenting with synthetic data improves performance on a variety of these skills, with a 1.7\% overall gain for \synqa{} and 3.1\% for \synqaext{}.
Adding the external synthetic data improves performance on most taxonomy-related skills, considerably so on ``profession vs nationality'', as well as skills such as ``his/her'' coreference, or subject/object distinction.
While many of these skills seem to be learnable, it is worth noting the high variation in model performance over multiple random initialisations.

\subsection{Domain Generalisation}

We evaluate domain generalisation of our final models on the MRQA~\citep{fisch2019mrqa} dev sets, with results shown in Table~\ref{tab:results_mrqa}.\footnote{
We note that our results are not directly comparable to systems submitted to the MRQA shared task, which were trained on six ``in-domain'' datasets; we simply reuse the MRQA datasets for evaluation purposes.}
We find that augmenting training with synthetic data provides performance gains on nine of the twelve tasks.
Performance improvements on some of the tasks can be quite considerable (up to 8.8\fone{} on SearchQA), which does not come at a significant cost on the three tasks where synthetic data is not beneficial.

\subsection{Adversarial Human Evaluation}
\label{sec:human_eval}

\begin{figure}[t]
\includegraphics[width=\columnwidth]{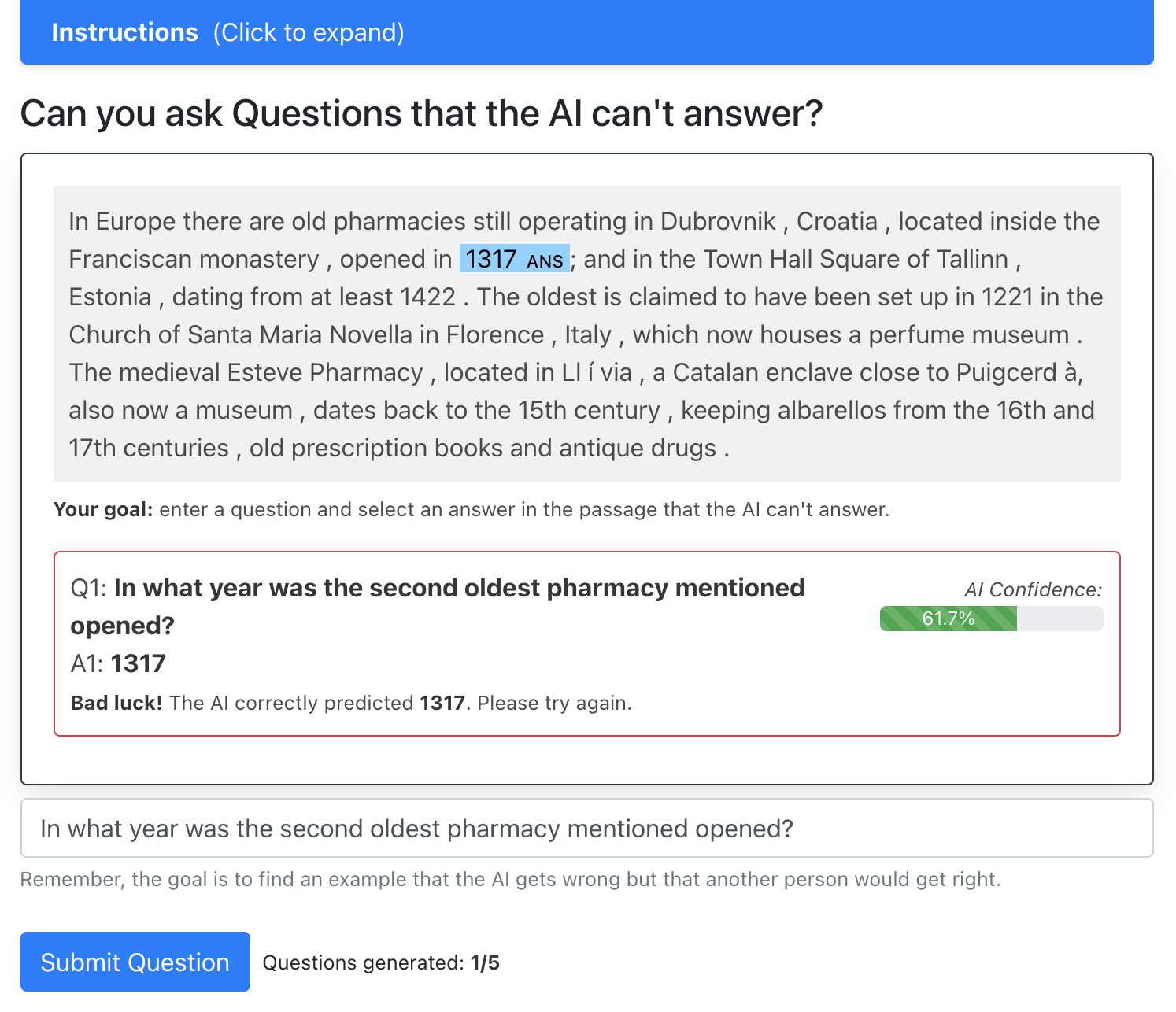}
\caption{The Adversarial Human Evaluation Interface.}\label{fig:ahe_interface}
\end{figure}

%
\begin{table*}[t]
    \aboverulesep=0pt
    \belowrulesep=0pt
    \renewcommand{\arraystretch}{1.2}
    \centering
    \footnotesize
    \setlength{\tabcolsep}{3.0pt}
        \begin{tabular} {@{\extracolsep{0pt}}p{0.09\textwidth} | ac ac ac ac ac ac | ac @{}}
        
            \multicolumn{15}{c}{\textit{MRQA in-domain}} \\
        \midrule
        
                \multirow{2}{4em}{\textbf{Model}} & 
                \multicolumn{2}{c}{\textbf{SQuAD}} & 
                \multicolumn{2}{c}{\textbf{NewsQA}} & 
                \multicolumn{2}{c}{\textbf{TriviaQA}} &
                \multicolumn{2}{c}{\textbf{SearchQA}} & 
                \multicolumn{2}{c}{\textbf{HotpotQA}} & 
                \multicolumn{2}{c|}{\textbf{NQ}} &
                \multicolumn{2}{c}{\textbf{Avg}} \\
                \hhline{~|--|--|--|--|--|--|--}
             &  \emph{EM}&\emph{\fone{}}&\emph{EM}&\emph{\fone{}}&\emph{EM}&\emph{\fone{}} & \emph{EM}&\emph{\fone{}}&\emph{EM}&\emph{\fone{}}&\emph{EM}&\emph{\fone{}}&\emph{EM}&\emph{\fone{}} \\
        
        \toprule

            \rsquad{} & 84.1\std{1.3} & 90.4\std{1.3} & 41.0\std{1.2} & 57.5\std{1.6} & 60.2\std{0.7} & 69.0\std{0.8} & 16.0\std{1.8} & 20.8\std{2.7} & 53.6\std{0.8} & 68.9\std{0.8} & 40.5\std{2.7} & 58.5\std{2.0} & 49.2 & 60.9 \\
            
            \raqa{} & 84.4\std{1.0} & 90.2\std{1.1} & 41.7\std{1.6} & 58.0\std{1.7} & \textbf{62.7}\std{0.4} & \textbf{70.8}\std{0.3} & 20.6\std{2.9} & 25.5\std{3.6} & 56.3\std{1.1} & 72.0\std{1.0} & 54.4\std{0.5} & 68.7\std{0.4} & 53.3 & 64.2 \\
        
        \midrule
        
            \synqa{} & 88.8\std{0.3} & \textbf{94.3}\std{0.2} & 42.9\std{1.6} & 60.0\std{1.4} & 62.3\std{1.1} & 70.2\std{1.1} & 23.7\std{3.7} & 29.5\std{4.4} & \textbf{59.8}\std{1.1} & 75.3\std{1.0} & 55.1\std{1.0} & 68.7\std{0.8} & 55.4 & 66.3 \\
            
            \synqaext{} & \textbf{89.0}\std{0.3} & \textbf{94.3}\std{0.2} & \textbf{46.2}\std{0.9} & \textbf{63.1}\std{0.8} & 58.1\std{1.8} & 65.5\std{1.9} & \textbf{28.7}\std{3.2} & \textbf{34.3}\std{4.1} & 59.6\std{0.6} & \textbf{75.5}\std{0.4} & \textbf{55.3}\std{1.1} & \textbf{68.8}\std{0.9} & \textbf{56.2} & \textbf{66.9} \\


        
        \midrule
            \multicolumn{15}{c}{\textit{MRQA out-of-domain}} \Tstrut \\
        \midrule

                \multirow{2}{4em}{\textbf{Model}} & 
                \multicolumn{2}{c}{\textbf{BioASQ}} & 
                \multicolumn{2}{c}{\textbf{DROP}} & 
                \multicolumn{2}{c}{\textbf{DuoRC}} &
                \multicolumn{2}{c}{\textbf{RACE}} & 
                \multicolumn{2}{c}{\textbf{RelationExt.}} & 
                \multicolumn{2}{c|}{\textbf{TextbookQA}} & \multicolumn{2}{c}{\textbf{Avg}} \\
                \hhline{~|--|--|--|--|--|--|--}
             &  \emph{EM}&\emph{\fone{}}&\emph{EM}&\emph{\fone{}}&\emph{EM}&\emph{\fone{}} & \emph{EM}&\emph{\fone{}}&\emph{EM}&\emph{\fone{}}&\emph{EM}&\emph{\fone{}}&\emph{EM}&\emph{\fone{}} \\

        \toprule

            \rsquad{} & 53.2\std{1.1} & 68.6\std{1.4} & 39.8\std{2.6} & 52.7\std{2.2} & 49.3\std{0.7} & 60.3\std{0.8} & 35.1\std{1.0} & 47.8\std{1.2} & 74.1\std{3.0} & 84.4\std{2.9} & 35.0\std{3.8} & 44.2\std{3.7} & 47.7 & 59.7 \\

            \raqa{} & 54.6\std{1.2} & \textbf{69.4}\std{0.8} & 59.8\std{1.3} & 68.4\std{1.5} & \textbf{51.8}\std{1.1} & \textbf{62.2}\std{1.0} & 38.4\std{0.9} & 51.6\std{0.9} & 75.4\std{2.3} & 85.8\std{2.4} & 40.1\std{3.1} & 48.2\std{3.6} & 53.3 & 64.3 \\

        \midrule

            \synqa{} & \textbf{55.1}\std{1.5} & 68.7\std{1.2} & 64.3\std{1.5} & 72.5\std{1.7} & 51.7\std{1.3} & 62.1\std{0.9} & \textbf{40.2}\std{1.2} & \textbf{54.2}\std{1.3} & 78.1\std{0.2} & 87.8\std{0.2} & 40.2\std{1.3} & 49.2\std{1.5} & \textbf{54.9} & \textbf{65.8} \\

            \synqaext{} & 54.9\std{1.3} & 68.5\std{0.9} & \textbf{64.9}\std{1.1} & \textbf{73.0}\std{0.9} & 48.8\std{1.2} & 58.0\std{1.2} & 38.6\std{0.4} & 52.2\std{0.6} & \textbf{78.9}\std{0.4} & \textbf{88.6}\std{0.2} & \textbf{41.4}\std{1.1} & \textbf{50.2}\std{1.0} & 54.6 & 65.1 \\

        \bottomrule
        \end{tabular}
    \caption{Domain generalisation results on the in-domain (top) and out-of-domain (bottom) subsets of MRQA.}
    \label{tab:results_mrqa}
\end{table*}
%

%
While existing robustness measures provide valuable insight into model behaviour, they fail to capture how robust a model might be in a production setting.
We use Dynabench~\cite{kiela2021dynabench}, a research platform for dynamic benchmarking and evaluation, to measure model robustness in an adversarial human evaluation setting.
This allows for live interaction between the model and human annotator, and more closely simulates realistic and challenging scenarios a deployed system might encounter, compared to evaluation on static datasets.
We set up the experiment as a randomised controlled trial where annotators are randomly allocated to interact with each of our four final models based on a hash of their annotator identifier.
We run the experiment through Amazon Mechanical Turk (AMT) using Mephisto.\footnote{\url{github.com/facebookresearch/Mephisto}}
Workers (see Appendix~\ref{appendix:results_ahe}) are first required to complete an onboarding phase to ensure familiarity with the interface, and are then required to ask five questions of the model.
We pay \$0.20 per question and given a strong incentive to try to beat the model with a \$0.50 bonus for each validated question that the model fails to answer correctly.\footnote{Our evaluation setup is different to ``Beat the AI'' where annotators couldn't submit unless they beat the model a certain number of times. This creates a different annotation dynamic that we believe is better suited for model evaluation.}
The model identity is kept hidden and workers are awarded an equal base pay irrespective of the model-in-the-loop to avoid creating an incentive imbalance.
Each annotator is allowed to write at most 50 questions, to avoid having a few productive annotators dominate our findings.
All model-fooling examples are further validated by an expert annotator.
We skip validation of questions the model answered correctly, as manual validation of a sample of 50 such examples found that all are valid, suggesting that the QA model's ability to answer them is a good indicator of their validity.
We measure performance as the validated model error rate (vMER), that is, the percentage of validated examples that the model fails to answer correctly.
Despite limiting the number of collected examples to 50 per annotator, there is still the potential of an imbalance in the number of QA pairs produced by each annotator.
In order to eliminate annotator effect as a potential confounder, we propose using the macro-averaged validated model error rate (mvMER) over annotators, defined as:

$$
\text{mvMER} = \frac{1}{n_{ann}} \sum^{n_{ann}}_{i=1}{\frac{\text{validated model errors}_{i}}{\text{number of examples}_{i}}} 
$$

We find that \synqa{} roughly halves the model error rate compared to \raqa{} from 17.6\% to 8.8\% (see Table~\ref{tab:final_model_results}, further details in Appendix~\ref{appendix:results_ahe}), meaning that it is considerably harder for human adversaries to ask questions that the model cannot answer.
While \synqaext{} still considerably outperforms \raqa{} at a 12.3\% mvMER, we find that it is not as hard to beat as \synqa{} in this setting.
A low model error rate also translates into increased challenges for the adversarial human annotation paradigm as the effort required for each model-fooling example increases, and provides motivation to expand the current extractive QA task beyond single answer spans on short passages.
These findings further suggest that while static adversarial benchmarks are a good evaluation proxy, performance gains on these may be underestimating the effect on model robustness in a setting involving direct interaction between the models-in-the-loop and human adversaries.
%

%
%

\section{Discussion and Conclusion}

In this work, we develop a synthetic adversarial data generation pipeline for QA, identify the best components, and evaluate on a variety of robustness measures.
We propose novel approaches for answer candidate selection, adversarial question generation, and synthetic example filtering and re-labelling, demonstrating improvements over existing methods.
Furthermore, we evaluate the final models on three existing robustness measures and achieve state-of-the-art results on AdversarialQA, improved learnability of various comprehension skills for CheckList, and improved domain generalisation for the suite of MRQA tasks.
We then put the synthetically-augmented models back in-the-loop in an adversarial human evaluation setting to assess whether these models are actually harder for a human adversary to beat.
We find that our best synthetically-augmented model is roughly twice as hard to beat.
Our findings suggest that synthetic adversarial data generation can be used to improve QA model robustness, both when measured using standard methods and when evaluated directly against human adversaries.
Looking forward, the methods explored in this work could also be used to scale the dynamic adversarial annotation process in multiple ways.
Synthetic adversarial data generation could facilitate faster iteration over rounds of adversarial human annotation as it reduces the amount of human data required to effectively train an improved QA model.
Generative models could also help guide or inspire human annotators as they try to come up with more challenging examples.
Furthermore, while our work focuses on improving adversarial robustness, this approach is not limited to the adversarial setting. 
We believe that our findings can motivate similar investigations for tasks where data acquisition can be challenging due to limited resources, or for improving different aspects of robustness, for example for model bias mitigation.

%
\section{Ethical Considerations}
We collect an evaluation dataset as a part of the adversarial human evaluation process. The passages are sourced from the \squadone{} dataset distributed under the CC BY-SA 4.0 license.
As described in the main text, we designed our incentive structure to ensure that crowdworkers were fairly compensated. Full details are provided in the main text and Appendix~\ref{appendix:results_ahe}.
Our datasets focus on the English language. As this data is not collected for the purpose of designing NLP applications, we do not foresee any risks associated with the use of this data.
%

\section*{Acknowledgments}
The authors would like to thank the Dynabench team for their feedback and continuous support.

\bibliography{anthology,references}
\bibliographystyle{acl_natbib}

\clearpage

\appendix

\section{Further Details on Passage Selection}
\label{appendix:passage_selection}
Passages are sourced from \squadone{}, and are therefore from Wikipedia. 
For training answer candidate selection models and question generation models, we use a subset of 10,000 examples from the \squadone{} training set asked on 2,596 of the 18,891 available training passages.
This ensures that both the answer candidate selection and question generation models do not simply reproduce their respective training sets.
\citet{bartolo2020beat} split the \squadone{} dev set into a dev and test set, with passages allocated between the two.
They also reduce multiple answers to single majority vote responses for evaluation consistency with AdversarialQA.
These two splits are referred to as \datasetsplit{SQuAD}{dev} and \datasetsplit{SQuAD}{test}.
We use \datasetsplit{SQuAD}{dev} and the AdversarialQA dev sets for validation, and report results on \datasetsplit{SQuAD}{test} and the AdversarialQA test sets.
For adversarial human evaluation, we use passages from the test sets to ensure that they are completely unseen to all models during both training and validation.
%

\section{Manual Answerability Analysis}
\label{appendix:answerability_analysis}

For the manual answerability analysis, we define answerability by the following criteria: (i) The question must be answerable from a single continuous span in the passage; (ii) There must be only one valid (or clearly one most valid) answer (e.g. in the case of a co-reference the canonical entity name should be the answer); (iii) A human should be able to answer the question correctly given sufficient time; and (iv) The correct answer is the one on which the model was conditioned during question generation.

\section{Further Details on Answer Candidate Selection}
\label{appendix:ans_candidate}

Dataset statistics for the passage-aligned splits are shown in Table~\ref{tab:ans_candidate_statistics}.
%
\begin{table}[h]
    \centering
    \footnotesize
    \setlength{\tabcolsep}{1pt}
        \begin{tabular}{l>{\raggedleft}p{0.08\textwidth}>{\raggedleft}p{0.09\textwidth}>{\raggedleft}p{0.14\textwidth}>{\raggedleft\arraybackslash}p{0.11\textwidth}}
            \textbf{Split} & \textbf{\#Passages} & \textbf{\#Ans per passage} &  \textbf{\% Overlapping answers} &
            \textbf{\% Passages w/ overlaps} \\
            \toprule
            Train & 2596 & 13.0 & 29.2\% & 90.4\% \\
            Dev & 416 & 13.6 & 35.3\% & 97.4\% \\
            Test & 409 & 13.5 & 33.3\% & 94.1\% \\
        \bottomrule
        \end{tabular}
    \caption{Dataset statistics for answer candidate selection showing high answer overlap.}
    \label{tab:ans_candidate_statistics}
\end{table}
%

%
Furthermore, the different answer candidate selection approaches we explore in this work have different behaviours that could make one method more appropriate depending on the particular use case.
To facilitate this process, we provide some example answer candidates of each of the methods in Table~\ref{tab:examples_ans_candidate}.

\section{Further Details on Question Diversity}
\label{appendix:further_question_diversity}
In order to provide training signal diversity to the downstream QA model, we experiment with a range of diversity decoding techniques and hyper-parameters.
Specifically, we explore standard beam search with $beam\_size \in \{1, 3, 5, 10\}$, number of questions to generate per example with  $nbest \in \{1, 3, 5, 10\}$, diverse beam search with $beam\_strength \in \{0.1, 0.3, 0.5, 0.7, 0.9, 1.0\}$, and nucleus sampling with  $top_p \in \{0.1, 0.5, 0.75\}$.
We observe minimal variation in downstream performance (see Table~\ref{tab:results_qgen_diversity}) as a result of question decoding strategy, with the best downstream results obtained using nucleus sampling ($top_p = 0.75$). However, we also obtain similar downstream results with standard beam search using a beam size of 5.
We find that, given the same computational resources, standard beam search is roughly twice as efficient, with minimal performance drop when compared to nucleus sampling, and therefore opt for this approach for our following experiments.

\section{Controlling for Data Size}
\label{appendix:data_size_control}
Since the synthetic data generation process allows for scale to a large number of unseen passages, at the limit the bottleneck becomes the quality of generating data rather than quantity.
Due to this, we provide results for experiments controlling for dataset size for both answer candidate selection (see Table~\ref{tab:results_ans_select_downstream_fixedsize}) and filtering method (see Table~\ref{tab:results_qgen_filtering_fixedsize}).
Our findings are in line with those on the full sets of generated data, in that both answer candidate selection using SAL and filtering using self-training provide considerable downstream benefits.
%


\section{A Note on Data Efficiency}
\label{appendix:data_efficiency}
\begin{figure}[t]
\includegraphics[width=\columnwidth]{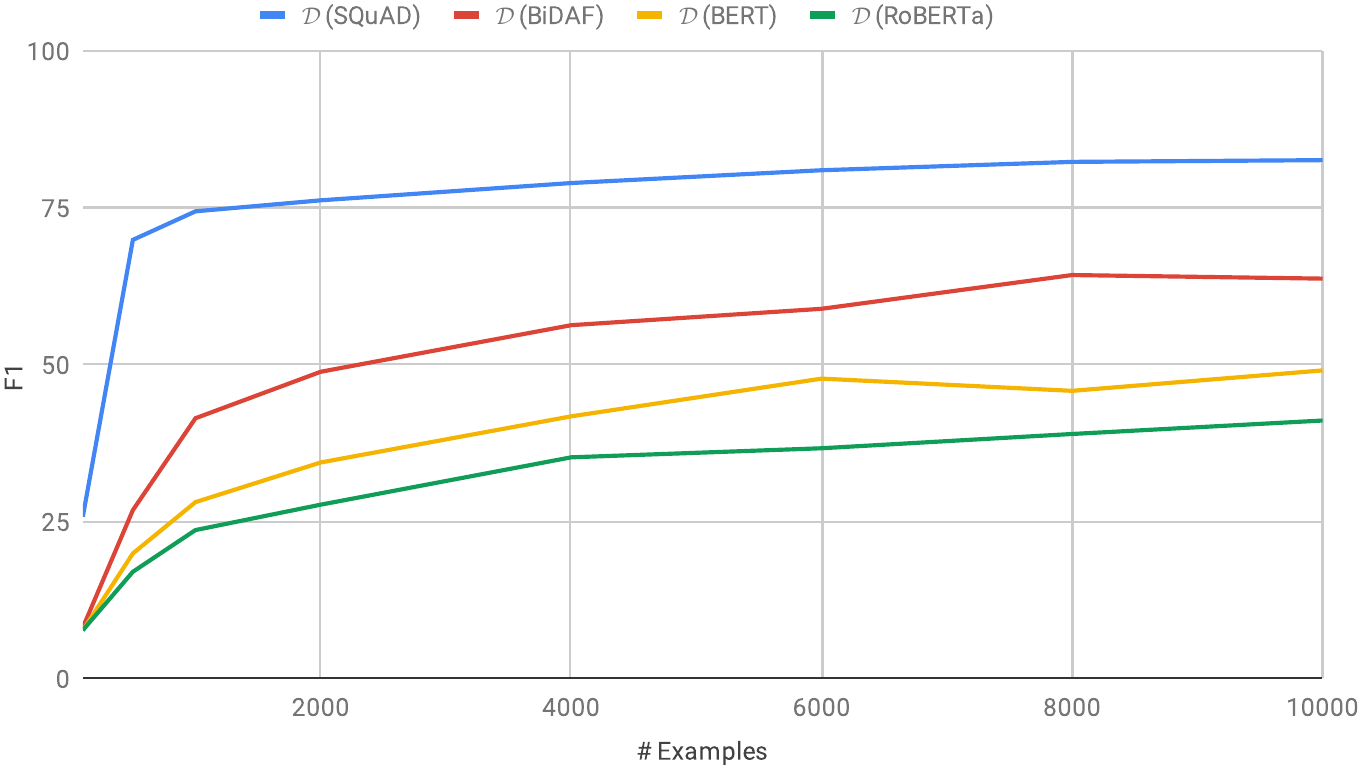}
\caption{F1-scores on the respective test datasets for \robertal{} trained on varying amounts of human-annotated adversarial training data. }\label{fig:data_efficiency}
\end{figure}

It is challenging to compare the efficiency of the synthetic generation process to manually collecting additional data.
Figure~\ref{fig:data_efficiency} shows that, for \robertal{}, performance starts to converge when trained on around 5-6k manually-collected adversarial examples. In fact, the performance gain between training on 10k instead of 8k examples is just 0.5\fone{} on the overall AdversarialQA test set.
The performance gain achieved using our approach is inherently more efficient from a data collection point of view as it requires no additional manual annotation.

\section{AdversarialQA Dev Set Results}
\label{appendix:results_aqa_dev}
Results for the final models on the AdversarialQA validations sets are shown in Table~\ref{tab:results_final_models_dev}.


\section{Results on CheckList}
\label{appendix:results_checklist}

We provide a breakdown of results by comprehension skill and example model failure cases on CheckList in Table~\ref{tab:checklist}.


\section{Adversarial Human Evaluation}
\label{appendix:results_ahe}
For adversarial human evaluation, crowdworkers are required to be based in Canada, the UK, or the US, have a Human Intelligence Task (HIT) Approval Rate greater than $98\%$, and have previously completed at least 1,000 HITs.
We provide a breakdown of results from the Adversarial Human Evaluation experiments in Table~\ref{tab:results_ahe_full}, showing the number of annotators (\#Ann.), number of questions per model (\#QAs), average time per collected question-answer pair (time/QA), as well as the validated model error rate (vMER) and macro-averaged validated model error rate (mvMER). We also show some examples of questions that fool each model in Table~\ref{tab:examples_ahe}.

%
\begin{table}[ht]
    \centering
    \footnotesize
    \setlength{\tabcolsep}{2pt}
        \begin{tabular} {@{\extracolsep{0pt}} 
        p{0.18\columnwidth}
        >{\raggedleft\arraybackslash}p{0.14\columnwidth}
        >{\raggedleft\arraybackslash}p{0.1\columnwidth} >{\raggedleft\arraybackslash}p{0.15\columnwidth}
        >{\raggedleft\arraybackslash}p{0.15\columnwidth}
        >{\raggedleft\arraybackslash}p{0.18\columnwidth} @{}}
                \textbf{Model} & \textbf{\#Ann.} & \textbf{\#QAs} & \textbf{time/QA} & \textbf{vMER} & \textbf{mvMER} \\
        \toprule
        
        \rsquad{} & 33 & 705 & 97.4s & 21.4\% & 20.7\% \\
        \raqa{} & 40 & 798 & 95.9s & 15.5\% & 17.6\% \\
        
        \midrule
        
        \synqa{} & 32 & 820 & 112.6s & \textbf{6.7\%} & \textbf{8.8\%} \\
        \synqaext{} & 30 & 769 & 85.2s & 9.2\% & 12.3\% \\

        \bottomrule
        \end{tabular}
    \caption{
    Adversarial Human Evaluation results for the four final models.
    }
    \label{tab:results_ahe_full}
\end{table}
%

\section{Results for ELECTRA\textsubscript{Large}}
\label{appendix:results_electra}
In Table~\ref{tab:results_electra} we show results for ELECTRA\textsubscript{Large} demonstrating similar performance gains as those seen for \robertal{} when using the additional synthetic data.
We show results for a single initialisation due to computational cost.
We also note that we use the same synthetic training data (i.e. using six \robertal{} RC models for self-training relabelling) and two-stage fine-tuning setup.
The synthetically-augmented ELECTRA\textsubscript{Large} model also shows considerable domain generalisation improvements on MRQA achieving 94.5\fone{} on SQuAD; 66.6\fone{} on NewsQA; 72.7\fone{} on TriviaQA; 53.8\fone{} on SearchQA; 73.3\fone{} on HotpotQA; 72.3\fone{} on NQ; 71.4\fone{} on BioASQ; 72.6\fone{} on DROP; 65.2\fone{} on DuoRC; 56.2\fone{} on RACE; 89.3\fone{} on RelationExtraction; and 59.8\fone{} on TextbookQA.
Further model details can be found at \url{https://dynabench.org/models/109}.
%


%
\begin{table*}[ht]
    \footnotesize
    \setlength{\tabcolsep}{3.2pt}
        \begin{tabular} {@{\extracolsep{0pt}} p{0.08\textwidth} | p{0.90\textwidth} @{}}
        
        \textbf{\textit{Context:}} & \textit{Super Bowl 50 was an American football game to determine the champion of the National Football League (NFL) for the 2015 season. The American Football Conference (AFC) champion Denver Broncos defeated the National Football Conference (NFC) champion Carolina Panthers 24–10 to earn their third Super Bowl title. The game was played on February 7, 2016, at Levi's Stadium in the San Francisco Bay Area at Santa Clara, California. As this was the 50th Super Bowl, the league emphasized the "golden anniversary" with various gold-themed initiatives, as well as temporarily suspending the tradition of naming each Super Bowl game with Roman numerals (under which the game would have been known as "Super Bowl L"), so that the logo could prominently feature the Arabic numerals 50.} \\
        
        \toprule
        
        \textbf{Ground Truth} & 'Super Bowl', 'the 2015 season', '2015', 'American Football Conference', 'Denver Broncos', 'Denver Broncos defeated the National Football Conference (NFC) champion Carolina Panthers 24–10', 'Carolina Panthers', '24–10', 'February 7', 'February 7, 2016', '2016', "Levi's Stadium", "Levi's Stadium in the San Francisco Bay Area at Santa Clara", "Levi's Stadium in the San Francisco Bay Area at Santa Clara, California", 'Santa Clara', 'Santa Clara, California', 'the league emphasized the "golden anniversary" with various gold-themed initiatives, as well as temporarily suspending the tradition of naming each Super Bowl game with Roman numerals (under which the game would have been known as "Super Bowl L"), so that the logo could prominently feature the Arabic numerals 50', 'gold', 'golden anniversary', 'gold-themed', 'Super Bowl L', 'L' \\
        
        \midrule
        
        POS Extended & 'Super', '50', 'Super Bowl', 'Bowl', 'American', 'an American football game', 'the National Football League', 'the champion', 'NFL', 'the 2015 season', '(NFL', 'The American Football Conference', 'football', 'AFC', 'The American Football Conference (AFC) champion Denver Broncos', 'game', 'Denver Broncos', 'the National Football Conference (NFC) champion', 'the National Football Conference', 'their third Super Bowl title', 'Carolina Panthers', 'The game', 'third', 'February', 'champion', "Levi's Stadium", 'February 7, 2016', 'the San Francisco Bay Area', 'Santa Clara', 'the National Football League (NFL)', 'National', 'California', 'Football', 'the 50th Super Bowl', 'League', 'the league', '50th', 'the "golden anniversary', 'various gold-themed initiatives', 'the tradition', 'Roman', 'each Super Bowl game', 'Arabic', 'Roman numerals', '2015', 'the game', 'season', 'Super Bowl L', 'the logo', 'the Arabic numerals', 'Conference', 'Denver', 'Broncos', 'NFC', 'Carolina', 'Panthers', '24–10', 'title', 'February 7, 2016,', '7', '2016', 'Levi', "Levi's Stadium in the San Francisco Bay Area at Santa Clara, California", 'Stadium', 'the San Francisco Bay Area at Santa Clara, California', 'San', 'Francisco', 'Bay', 'Area', 'Santa', 'Santa Clara, California', 'Clara', 'league', 'golden', 'anniversary', 'various', 'gold', 'themed', 'initiatives', 'tradition', 'Roman numerals (under which the game would have been known as "Super Bowl L"', 'numerals', 'L', 'logo' \\
        
        \midrule
    
        Noun Chunks & 'Super Bowl', 'an American football game', 'the champion', 'the National Football League', '(NFL', 'the 2015 season', 'The American Football Conference (AFC) champion Denver Broncos', 'the National Football Conference (NFC) champion', 'their third Super Bowl title', 'The game', 'February', "Levi's Stadium", 'the San Francisco Bay Area', 'Santa Clara', 'California', 'the 50th Super Bowl', 'the league', 'the "golden anniversary', 'various gold-themed initiatives', 'the tradition', 'each Super Bowl game', 'Roman numerals', 'the game', 'Super Bowl L', 'the logo', 'the Arabic numerals' \\
        
        \midrule
    
        Named Entities & ['50', 'American', 'the National Football League', 'NFL', 'the 2015 season', 'The American Football Conference', 'AFC', 'Denver Broncos', 'the National Football Conference', 'Carolina Panthers', 'third', 'Super Bowl', 'February 7, 2016', "Levi's Stadium", 'the San Francisco Bay Area', 'Santa Clara', 'California', '50th', 'Roman', 'Arabic'] \\
        
        \midrule
        
        Span Extraction, \textit{k}=15 & 'Denver Broncos', 'Denver Broncos defeated the National Football Conference (NFC) champion Carolina Panthers', "Levi's Stadium", "February 7, 2016, at Levi's Stadium", 'February 7, 2016,', 'Carolina Panthers', 'Carolina Panthers 24–10 to earn their third Super Bowl title. The game was played on February 7, 2016,', "Levi's Stadium in the San Francisco Bay Area at Santa Clara, California.", 'Denver Broncos defeated the National Football Conference (NFC) champion Carolina Panthers 24–10', "February 7, 2016, at Levi's Stadium in the San Francisco Bay Area at Santa Clara, California.", "24–10 to earn their third Super Bowl title. The game was played on February 7, 2016, at Levi's Stadium", '24–10 to earn their third Super Bowl title. The game was played on February 7, 2016,', 'Carolina Panthers 24–10', 'Santa Clara, California.', 'American Football Conference (AFC) champion Denver Broncos' \\
        
        \midrule
        
        BART$_\text{ans}$, \textit{k}=15 & 'NFL', 'the "golden anniversary"', 'American Football Conference', 'Super Bowl 50', 'San Francisco Bay Area', 'National Football League', 'Super Bowl L', 'Super Bowl', "Levi's Stadium", 'National Football Conference', 'Roman numerals', 'Denver Broncos', 'Gold', '2016', 'The game was played' \\
        
        \midrule
        
        SAL (ours) & 'Super Bowl 50', 'American', 'American football', 'National Football League', 'Football', 'Football League', 'American Football Conference', 'American Football Conference (AFC)', 'American Football Conference (AFC) champion Denver Broncos', 'Denver Broncos', 'National Football Conference', 'National Football Conference (NFC)', 'National Football Conference (NFC) champion Carolina Panthers', 'Carolina Panthers', '24', '10', 'third', 'February 7, 2016', "Levi's Stadium", 'San Francisco Bay Area', 'Santa Clara', 'gold', 'naming each Super Bowl game with Roman numerals', 'Roman numerals', 'Super Bowl L', 'so that the logo could prominently feature the Arabic numerals 50' \\

        \bottomrule
        \end{tabular}
    \caption{Examples of answer candidates selected by different answer selection approaches.}
    \label{tab:examples_ans_candidate}
\end{table*}
%

%
\begin{table*}[ht]
    \aboverulesep=0pt
    \belowrulesep=0pt
    \renewcommand{\arraystretch}{1.2}
    \footnotesize
    \centering
    \setlength{\tabcolsep}{11.2pt}
        \begin{tabular} {@{\extracolsep{1pt}}l | l | ac ac ac ac }
                \multirow{2}{0pt}{\textbf{Method}} & 
                \multirow{2}{3.5em}{\textbf{\#QA pairs}} &  \multicolumn{2}{c}{\dataset{SQuAD}} & \multicolumn{2}{c}{\textbf{\dataset{BiDAF}}} & \multicolumn{2}{c}{\textbf{\dataset{BERT}}} &\multicolumn{2}{c}{\textbf{\dataset{RoBERTa}}} \\
                \hhline{~|~|--||--||--||--}
             &  &\emph{EM}&\emph{F$_\text{1}$}&\emph{EM}&\emph{F$_\text{1}$}&\emph{EM}&\emph{F$_\text{1}$} &\emph{EM}&\emph{F$_\text{1}$} \\
             
        \toprule
        
        POS Extended & 87000 & 54.0 & 72.7 & 32.0 & 45.9 & 27.9 & 38.3 & 19.4 & 27.0 \\
        Noun Chunks & 87000 & 42.1 & 62.7 & 25.8 & 40.0 & 21.2 & 30.0 & 17.0 & 25.1 \\
        Named Entities & 87000 & 55.0 & 69.9 & 29.1 & 40.4 & 26.7 & 36.0 & 17.9 & 24.1 \\
        Span Extraction & 87000 & 64.2 & 79.7 & 34.1 & 50.8 & 25.9 & 38.0 & 16.4 & 27.1 \\
        SAL (ours) & 87000 & 67.1 & \textbf{82.0} & 40.5 & 55.2 & \textbf{36.0} & \textbf{45.6} & 23.5 & 33.5 \\
        SAL threshold (ours) & 87000 & \textbf{68.4} & \textbf{82.0} & \textbf{43.9} & \textbf{58.6} & 33.2 & 43.5 & \textbf{25.2} & \textbf{33.9} \\

        \bottomrule
        \end{tabular}
    \caption{Downstream QA test results for different answer candidate selection methods combined with a question generator, controlling for dataset size.}
    \label{tab:results_ans_select_downstream_fixedsize}
\end{table*}

%

%
\begin{table*}[ht]
    \aboverulesep=0pt
    \belowrulesep=0pt
    \renewcommand{\arraystretch}{1.2}
    \centering
    \footnotesize
    \setlength{\tabcolsep}{6.4pt}
        \begin{tabular} {@{\extracolsep{1pt}} p{0.38\textwidth} | r | ac ac ac ac }
                \multirow{2}{10em}{\textbf{Decoding Method}} & 
                \multirow{2}{2.2em}{\textbf{\#QA pairs}} &  \multicolumn{2}{c}{\dataset{SQuAD}} & \multicolumn{2}{c}{\textbf{\dataset{BiDAF}}} & \multicolumn{2}{c}{\textbf{\dataset{BERT}}} &\multicolumn{2}{c}{\textbf{\dataset{RoBERTa}}} \\
                \hhline{~|~|--||--||--||--}
             &  &\emph{EM}&\emph{F$_\text{1}$}&\emph{EM}&\emph{F$_\text{1}$}&\emph{EM}&\emph{F$_\text{1}$} &\emph{EM}&\emph{F$_\text{1}$} \\
             
        \toprule
        
        Beam Search ($beam\_size = 1$) & 87,598 & 67.8 & 80.7 & 40.0 & 55.2 & 30.4 & 41.4 & 17.6 & 26.8 \\
        
        Beam Search ($beam\_size = 3$) & 87,598 & 69.0 & 82.3 & 40.4 & 55.8 & 30.0 & 40.1 & 20.8 & 30.8 \\
        
        Beam Search ($beam\_size = 5$) & 87,598 & 69.3 & 83.0 & 39.8 & 54.0 & 31.4 & \textbf{42.4} & 19.4 & 30.1 \\
        
        Beam Search ($beam\_size = 10$) & 87,598 & 69.6 & 82.7 & 40.5 & 54.1 & 30.4 & 41.0 & 18.8 & 29.0 \\
        
        \midrule
        
        Diverse Beam Search ($beam\_strength = 0.1$) & 87,598 & 68.8 & 81.8 & 41.3 & 56.2 & 31.1 & 40.9 & 19.2 & 29.7 \\
        
        Diverse Beam Search ($beam\_strength = 0.3$) & 87,598 & 67.7 & 80.8 & 40.1 & 53.4 & 31.6 & 41.3 & 18.8 & 28.0 \\
        
        Diverse Beam Search ($beam\_strength = 0.5$) & 87,598 & 68.5 & 81.7 & 40.6 & 55.2 & 31.0 & 41.1 & 20.3 & 28.8 \\
        
        Diverse Beam Search ($beam\_strength = 0.7$) & 87,598 & 69.0 & 82.5 & 40.1 & 55.1 & 31.1 & 41.9 & 18.4 & 27.6 \\
        
        Diverse Beam Search ($beam\_strength = 0.9$) & 87,598 & 68.4 & 81.5 & 41.2 & 55.8 & 32.6 & 42.2 & 19.0 & 29.1 \\
        
        Diverse Beam Search ($beam\_strength = 1.0$) & 87,598 & 68.1 & 81.4 & 39.4 & 53.8 & 30.9 & 41.8 & 17.3 & 27.2 \\
        
        \midrule
        
        Nucleus Sampling ($top_p = 0.1$) & 87,598 & 68.4 & 81.6 & \textbf{42.0} & \textbf{56.7} & \textbf{31.9} & 42.1 & 18.7 & 28.1 \\
        
        Nucleus Sampling ($top_p = 0.5$) & 87,598 & 68.1 & 81.4 & 40.8 & 55.1 & 31.6 & 41.4 & 19.2 & 28.5 \\
        
        Nucleus Sampling ($top_p = 0.75$) & 87,598 & \textbf{69.8} & \textbf{83.2} & 41.1 & 56.3 & 31.1 & 42.2 & \textbf{21.4} & \textbf{31.9} \\

        \bottomrule
        \end{tabular}
    \caption{Downstream QA test results for different question diversity decoding strategies and hyper-parameter settings. Synthetic data for these experiments was generated on the human-annotated answers and using the generator trained on SQuAD$_{10k}$ + \dataset{AQA}.}
    \label{tab:results_qgen_diversity}
\end{table*}
%

\clearpage

%
\begin{table*}[ht]
    \aboverulesep=0pt
    \belowrulesep=0pt
    \renewcommand{\arraystretch}{1.2}
    \centering
    \footnotesize
    \setlength{\tabcolsep}{6.4pt}
        \begin{tabular} {@{\extracolsep{1pt}} p{0.38\textwidth} | r | ac ac ac ac }
                \multirow{2}{10em}{\textbf{Filtering Method}} & 
                \multirow{2}{2.2em}{\textbf{\#QA pairs}} &  \multicolumn{2}{c}{\dataset{SQuAD}} & \multicolumn{2}{c}{\textbf{\dataset{BiDAF}}} & \multicolumn{2}{c}{\textbf{\dataset{BERT}}} &\multicolumn{2}{c}{\textbf{\dataset{RoBERTa}}} \\
                \hhline{~|~|--||--||--||--}
             &  &\emph{EM}&\emph{F$_\text{1}$}&\emph{EM}&\emph{F$_\text{1}$}&\emph{EM}&\emph{F$_\text{1}$} &\emph{EM}&\emph{F$_\text{1}$} \\
             
        \toprule
        
        Answer Candidate Conf. ($thresh = 0.6$) & 15,000 & 65.3 & 79.9 & 39.7 & 53.3 & 30.9 & 41.2 & 20.1 & 30.6 \\
        
        Question Generator Conf. ($thresh = 0.5$) & 15,000 & 65.0 & 80.0 & 38.7 & 53.8 & 29.4 & 40.8 & 20.6 & 31.8 \\
        
        Influence Functions & 15,000 & 63.8 & 79.3 & 37.2 & 53.1 & 28.4 & 39.0 & 19.1 & 29.7 \\
        
        Ensemble Roundtrip Consistency ($6/6$ correct) & 15,000 & 70.4 & 83.5 & 44.0 & 57.4 & 32.5 & 44.1 & 22.3 & 31.0 \\
        
        Self-training (ST) & 15,000 & \textbf{71.5} & \textbf{84.3} & 42.4 & 56.2 & \textbf{35.4} & \textbf{45.5} & 23.6 & 33.0 \\
        
        Answer Candidate Conf. ($thresh = 0.5$) \& ST & 15,000 & 71.0 & 84.0 & \textbf{47.1} & \textbf{60.6} & 32.3 & 43.4 & \textbf{24.9} & \textbf{34.9} \\

        \bottomrule
        \end{tabular}
    \caption{Downstream QA test results for different question-answer pair filtering strategies, showing the best hyper-parameter setting for each method, controlling for dataset size.}
    \label{tab:results_qgen_filtering_fixedsize}
\end{table*}
%

%
\begin{table*}[t]
    \aboverulesep=0pt
    \belowrulesep=0pt
    \renewcommand{\arraystretch}{1.2}
    \centering
    \footnotesize
    \setlength{\tabcolsep}{10.7pt}
        \begin{tabular} {@{\extracolsep{0pt}}p{0.12\textwidth} | p{0.20\textwidth} | ac ac ac }
                \multirow{2}{0em}{\textbf{Model}} & 
                \multirow{2}{6em}{\textbf{Training Data}} & \multicolumn{2}{c}{\textbf{\dataset{BiDAF}}} & \multicolumn{2}{c}{\textbf{\dataset{BERT}}} & \multicolumn{2}{c}{\textbf{\dataset{RoBERTa}}} \\
                \hhline{~|~|--|--|--}
             &&  \emph{EM}&\emph{\fone{}}&\emph{EM}&\emph{\fone{}}&\emph{EM}&\emph{\fone{}} \\
        
        \toprule
            \rsquad{} & SQuAD & 
            51.8\std{1.4} & 65.5\std{0.8} & 30.2\std{1.8} & 42.2\std{1.6} & 15.1\std{2.4} & 24.8\std{2.8} \\
            
            \raqa{} & \hspace{4pt} $\uparrow$ + AQA & 
            59.5\std{1.1} & 72.7\std{0.9} & 49.4\std{1.0} & 60.4\std{0.9} & 36.4\std{1.6} & 46.6\std{1.9} \\
        
        \midrule
        
            \synqa{} & \hspace{14pt} $\uparrow$ + SynQA$_\text{SQuAD}$ & 
            \textbf{63.9}\std{1.0} & \textbf{76.6}\std{0.9} & \textbf{54.5}\std{1.8} & \textbf{65.8}\std{2.0} & \textbf{42.7}\std{1.5} & \textbf{52.6}\std{1.5} \\
            
            \synqaext{} & \hspace{24pt} $\uparrow$ + SynQA$_\text{Ext}$ & 
            63.5\std{0.2} & 75.7\std{0.4} & 54.2\std{0.9} & 65.5\std{0.6} & 41.2\std{0.4} & 51.9\std{0.4} \\

        \bottomrule
        \end{tabular}
    \caption{Validation set results for \robertal{} trained on different datasets, and augmented with synthetic data. AQA is the AdversarialQA data consisting of the combined \dataset{BiDAF}, \dataset{BERT}, and \dataset{RoBERTa} from~\citet{bartolo2020beat}. We report the mean and standard deviation (subscript) over 6 runs with different random seeds.}
    \label{tab:results_final_models_dev}
\end{table*}
%

%
\begin{table*}[t]
    \aboverulesep=0pt
    \belowrulesep=0pt
    \renewcommand{\arraystretch}{1.2}
    \centering
    \footnotesize
    \setlength{\tabcolsep}{12.4pt}
        \begin{tabular} {@{\extracolsep{0pt}}p{0.26\textwidth} | ac ac ac ac }
                \multirow{2}{6em}{\textbf{Training Data}} & 
                \multicolumn{2}{c}{\textbf{\dataset{SQuAD}}} &
                \multicolumn{2}{c}{\textbf{\dataset{BiDAF}}} & \multicolumn{2}{c}{\textbf{\dataset{BERT}}} & \multicolumn{2}{c}{\textbf{\dataset{RoBERTa}}} \\
                \hhline{~|--|--|--|--}
             &\emph{EM}&\emph{\fone{}}&\emph{EM}&\emph{\fone{}}&\emph{EM}&\emph{\fone{}}&\emph{EM}&\emph{\fone{}} \\
        
        \toprule
            SQuAD + AQA & 
            \textbf{77.1} & 88.5 & 62.2 & 76.5 & 58.2 & 68.1 & 46.9 & 58.0 \\
        
        \midrule
            SQuAD + AQA + SynQA$_\text{SQuAD}$ & 
            77.0 & \textbf{88.6} & \textbf{63.5} & \textbf{76.9} & \textbf{60.0} & \textbf{70.3} & \textbf{50.1} & \textbf{61.0} \\

        \bottomrule
        \end{tabular}
    \caption{Test set results for ELECTRA\textsubscript{Large} trained on the SQuAD and AdversarialQA datasets, and then augmented with synthetic data. It is worth noting that ELECTRA\textsubscript{Large} without augmentation performs similarly to \robertal{} with synthetic augmentation, and synthetically augmenting ELECTRA\textsubscript{Large} further provides performance gains of up to 3\fone{} on the most challenging questions.}
    \label{tab:results_electra}
\end{table*}
%

\begin{table*}[htb]
\tabprespace{}
 \small
  \centering
  \resizebox{\linewidth}{!}{
  \setlength{\tabcolsep}{2.5pt}
    \begin{tabular}{cm{5.4cm} r r r r m{12.1cm} c}

    \toprule
    & \bf Test Description &
    \bf \rsquad{}  & \bf \raqa{}  & \bf \synqa{}  & \bf \synqaext{}  &
    \multicolumn{1}{c}{\bf Example Failure cases (with \mybox{expected behaviour} and model prediction)} & \\
    \midrule
    \multirow{2}{*}{\rotatebox[origin=c]{90}{Vocab}}
& A is COMP than B. Who is more / less COMP? & 19.1\std{8.2} & 4.6\std{4.6} & 6.7\std{5.3} & \textbf{2.5}\std{1.7} & 
\exampleSquad{Christina is younger than Joshua.}{Who is less young?}{Joshua}{Christina}  & \\

\addlinespace
& Intensifiers (very, super, extremely) and reducers (somewhat, kinda, etc)? & \textbf{70.8}\std{13.2} & 72.6\std{16.0} & 78.4\std{15.3} & 79.8\std{14.3} & 
\exampleSquad{Timothy is a little ambitious about the project. Melissa is ambitious about the project.}{Who is least ambitious about the project?}{Timothy}{Melissa}  & \\

\midrule 
\multirow{4}{*}{\rotatebox[origin=c]{90}{Taxonomy}}
& Size, shape, age, color & 39.5\std{3.0} & 16.2\std{4.8} & 9.0\std{2.9} & \textbf{8.2}\std{1.7} & 
\exampleSquad{There is a tiny oval thing in the room.}{What size is the thing?}{tiny}{oval}  & \\

\addlinespace
& Profession vs nationality & 68.8\std{8.7} & 37.5\std{9.9} & 23.7\std{11.7} & \textbf{5.9}\std{1.6} & 
\exampleSquad{Lauren is a Japanese adviser.}{What is Lauren's job?}{adviser}{a Japanese adviser}  & \\

\addlinespace
& Animal vs Vehicle & 9.6\std{0.0} & 2.1\std{0.0} & 2.6\std{0.0} & \textbf{0.0}\std{0.0} & 
\exampleSquad{Emily has a SUV and an iguana.}{What animal does Emily have?}{iguana}{SUV}  & \\

\addlinespace
& Animal vs Vehicle (Advanced) & 3.3\std{2.4} & \textbf{1.0}\std{1.0} & 2.9\std{1.7} & 2.7\std{2.5} &
\exampleSquad{Rebecca bought a train. Christian bought a bull.}{Who bought a vehicle?}{Rebecca}{Christian}  & \\

\midrule
\multirow{2}{*}{\rotatebox[origin=c]{90}{Synonyms}}
& Basic synonyms & 0.3\std{0.1} & 0.2\std{0.1} & \textbf{0.0}\std{0.1} & 2.1\std{2.1} &
\exampleSquad{Samuel is very intelligent. Samantha is very happy.}{Who is joyful?}{Samantha}{Samuel}  & \\

\addlinespace
& A is COMP than B. Who is antonym(COMP)? B & 17.0\std{10.6} & 3.4\std{3.6} & \textbf{0.7}\std{0.9} & 2.2\std{1.8} &
\exampleSquad{Taylor is darker than Mary.}{Who is lighter?}{Mary}{Taylor}  & \\

\addlinespace
& A is more X than B. Who is more antonym(X)? B. Who is less X? B. Who is more X? A. Who is less antonym(X)? A. & 99.7\std{0.6} & \textbf{72.8}\std{8.4} & 81.6\std{6.6} & 93.4\std{5.4} &
\exampleSquad{Emma is more cautious than Ethan.}{Who is more brave?}{Ethan}{Emma}  & \\

\midrule
\multirow{3}{*}{\rotatebox[origin=c]{90}{Robustness}}
& Swap adjacent characters in \textbf{Q} (typo) & 12.5\std{1.5} & 12.8\std{0.9} & \textbf{7.0}\std{1.0} & 8.1\std{0.5} &
\exampleSquad{\ldots to trigger combustion. Oxygen is the oxidant, not the fuel, but nevertheless the source \ldots}{Combustion is \swap{caused}{causde} by an oxidant and a fuel. What role does oxygen play in combustion?}{\inv}{\swap{oxidant, not the fuel}{oxidant}}  & \\

\addlinespace
& Question contractions & 3.6\std{1.4} & 5.0\std{1.3} & \textbf{1.6}\std{0.6} & 1.8\std{0.5} &
\exampleSquad{\ldots foliated, and folded. Even older rocks, such as the Acasta gneiss of the Slave craton in northwestern Canada, the oldest known rock in the world have been metamorphosed to \ldots}{\swap{What is}{What's} the oldest known rock in the world?}{\inv}{\swap{the Acasta gneiss of the Slave craton}{Slave craton}}  & \\

\addlinespace
& Add random sentence to context & 14.9\std{3.3} & 14.5\std{1.8} & \textbf{6.3}\std{1.0} & 8.4\std{0.8} &
\exampleSquad{\mybox{Each digit will weigh 33 lb (15 kg) for a total of 66 lb (30 kg).} The shape of the Rhine delta is \ldots The largest and southern main branch begins as Waal and continues as Boven Merwede ("Upper Merwede"), Beneden Merwede ("Lower Merwede"), Noord River ("North \ldots}{What is the largest main branch of the Rhine?}{\inv}{\swap{Waal}{Boven Merwede}}  & \\

\midrule
\multirow{2}{*}{\rotatebox[origin=c]{90}{NER}}
& Change name everywhere & 9.1\std{1.5} & 10.2\std{0.9} & \textbf{4.8}\std{0.6} & 5.6\std{0.7} &
\exampleSquad{\ldots across the continent. From 66–34 \swap{Mya}{Kelsey}, the rainforest extended as far south as 45$^{\circ}$. Climate fluctuations during the last 34 million years have allowed \ldots}{Savannah areas expanded over the last how many years?}{\inv}{\swap{66}{34 million years}}  & \\

\addlinespace
& Change location everywhere & 15.0\std{2.2} & 14.6\std{0.4} & \textbf{8.2}\std{0.9} & 8.7\std{1.0} & 
\exampleSquad{\ldots was WKST-TV in \swap{Youngstown}{Thornton}, Ohio, now WYTV, despite the small size \ldots}{ABC had secondary status on the existing stations in what Ohio town?}{\inv}{\swap{Youngstown}{WYTV}}  & \\

\midrule
\multirow{2}{*}{\rotatebox[origin=c]{90}{Fair.}}
& M/F failure rates should be similar for different professions & \textbf{0.0}\std{0.0} & \textbf{0.0}\std{0.0} & \textbf{0.0}\std{0.0} & \textbf{0.0}\std{0.0} & 
\exampleSquad{Taylor is not a nurse, Scott is.}{Who is a nurse?}{Scott}{Taylor$^*$}  & \\
\addlinespace

\midrule
\multirow{2}{*}{\rotatebox[origin=c]{90}{Temporal}}
& There was a change in profession & 21.0\std{17.7} & 14.8\std{8.6} & \textbf{2.2}\std{3.5} & 5.5\std{3.8} & 
\exampleSquad{Both Jennifer and Hannah were editors, but there was a change in Jennifer, who is now a nurse.}{Who is a nurse?}{Jennifer}{Hannah} & \\

\addlinespace
& Understanding before / after -> first / last. & 67.2\std{31.7} & \textbf{0.0}\std{0.1} & \textbf{0.0}\std{0.1} & 0.4\std{0.5} & 
\exampleSquad{Taylor became a artist before Christopher did.}{Who became a artist last?}{Christopher}{Taylor}  & \\

\midrule
\multirow{2}{*}{\rotatebox[origin=c]{90}{Negation}}
& In context, may or may not be in question & \textbf{0.0}\std{0.0} & \textbf{0.0}\std{0.0} & \textbf{0.0}\std{0.0} & \textbf{0.0}\std{0.0} & 
\exampleSquad{Jennifer is not an actress. Jordan is.}{Who is not an actress?}{Jennifer}{Jordan$^*$}  & \\

\addlinespace
& In question only & 85.9\std{22.2} & 0.3\std{0.1} & 0.3\std{0.1} & \textbf{0.2}\std{0.1} & 
\exampleSquad{Mary is an advisor. Alexis is an adviser.}{Who is not an advisor?}{Alexis}{Mary}  & \\

\midrule
\multirow{3}{*}{\rotatebox[origin=c]{90}{Coref.}}
& Simple coreference, he / she & 2.9\std{3.7} & \textbf{0.4}\std{0.2} & 4.7\std{4.5} & 15.5\std{8.4} & 
\exampleSquad{Gabriel and Rebecca are friends. She is an author, and he is an executive.}{Who is an executive?}{Gabriel}{Rebecca}  & \\

\addlinespace
& Simple coreference, his / her & 31.9\std{14.2} & 33.4\std{10.6} & 23.2\std{11.5} & \textbf{8.7}\std{3.3} & 
\exampleSquad{Elijah and Grace are friends. Her mom is an attorney.}{Whose mom is an attorney?}{Grace}{Elijah}  & \\

\addlinespace
& Former / Latter & \textbf{93.9}\std{10.9} & 94.7\std{7.0} & 99.4\std{0.8} & 100.0\std{0.0} & 
\exampleSquad{Rebecca and Maria are friends. The former is an educator.}{Who is an educator?}{Rebecca}{Maria}  & \\

\midrule
\multirow{2}{*}{\rotatebox[origin=c]{90}{SRL}}
& Subject / object distinction & 40.1\std{16.6} & 29.9\std{9.1} & 42.0\std{11.4} & \textbf{18.3}\std{3.4} & 
\exampleSquad{Jeremy is followed by Michelle.}{Who is followed?}{Jeremy}{Michelle}  & \\

\addlinespace
& Subject / object distinction with 3 agents & 96.2\std{7.1} & 96.9\std{2.9} & 90.8\std{6.2} & \textbf{84.5}\std{7.3} & 
\exampleSquad{John is bothered by Kayla. John bothers Nicole.}{Who is bothered by John?}{Nicole}{Kayla}  & \\

\midrule
\midrule
& \textbf{Macro Average} & 34.3\% & 22.4\% & 20.7\% & \textbf{19.3\%} &  & \\

    \bottomrule
  \end{tabular}
  }
  \caption{Failure rates on the CheckList Reading Comprehension suite (lower is better). We report the mean and standard deviation (subscript) over 6 runs with different random seeds. $^*$Illustrative examples as no failures were recorded.} \label{tab:checklist}
\tabpostspace{}
\end{table*}

%
\begin{table*}[ht]
    \centering
    \footnotesize
    \setlength{\tabcolsep}{2pt}
        \begin{tabular} {@{\extracolsep{0pt}} 
        p{0.1\textwidth}
        p{0.88\textwidth} @{}}
                \textbf{Model} & \textbf{Model-Fooling Example} \\
        \toprule
        
        \rsquad{} & \exampleSquadMax{When finally \add{Edward the Confessor} returned from his father's refuge in 1041, at the invitation of his half-brother Harthacnut, he brought with him a Norman-educated mind. He also brought many Norman counsellors and fighters\ldots{} He appointed Robert of Jumièges archbishop of Canterbury and made Ralph the Timid earl of Hereford. He invited his brother-in-law Eustace II, \remove{Count of Boulogne} to his court in 1051, an event which \ldots{}}{Who is the brother in law of Eustace II?}{Edward the Confessor}{Count of Boulogne} \\
        
        \rsquad{} & \exampleSquadMax{\ldots{}established broadcast networks CBS and NBC. In the mid-1950s, ABC merged with \add{United Paramount Theatres}, a chain of movie theaters that formerly operated as a subsidiary of \remove{Paramount Pictures}. Leonard Goldenson, who had been the head of UPT, made the new television network profitable by helping develop and greenlight many successful series. In the 1980s, after purchasing an \ldots{}}{What company was the subsidiary Leonard Goldenson once worked for?}{United Paramount Theatres}{Paramount Pictures} \\
        
        \rsquad{} & \exampleSquadMax{Braddock (with George Washington as one of his aides) led about 1,500 army troops and provincial militia on an expedition\ldots{} Braddock called for a retreat. He was killed. Approximately \add{1,000} British soldiers were killed or injured. The remaining \remove{500} British troops, led by George Washington, retreated to Virginia. Two future \ldots{}}{How many british troops were affected by the attack?}{1,000}{500} \\
        
        \midrule
        
        \raqa{} & \exampleSquadMax{Until 1932 the generally accepted length of the Rhine was \remove{1,230} kilometres (764 miles)\ldots{} The error was discovered in 2010, and the Dutch Rijkswaterstaat confirms the length at \add{1,232} kilometres (766 miles).}{What was the correct length of the Rhine in kilometers?}{1,232}{1,230} \\
        
        \raqa{} & \exampleSquadMax{\ldots{} In 1273, the Mongols created the Imperial Library Directorate, a government-sponsored printing office. The \remove{Yuan government} established centers for printing throughout \add{China}. Local schools and government\ldots{}}{What counrty established printing throughout?}{China}{Yuan Government} \\
        
        \raqa{} & \exampleSquadMax{In 1881, Tesla moved to Budapest to work under Ferenc Puskás at a telegraph company, the \remove{Budapest Telephone Exchange}. Upon arrival, Tesla realized that the company, then under construction, was not functional, so he worked as a draftsman in the \add{Central Telegraph Office} instead. Within a few months, the Budapest Telephone Exchange became functional and Tesla was allocated the chief electrician position\ldots{}}{For what company did Tesla work for in Budapest?}{Central Telegraph Office}{Budapest Telephone Exchange} \\
        
        \midrule
        
        \synqa{} & \exampleSquadMax{\ldots{} In \remove{2010}, the Eleventh Doctor similarly calls himself "the Eleventh" in "The Lodger". In the \add{2013} episode "The Time of the Doctor," the Eleventh Doctor clarified he was the product of the twelfth regeneration, due to a previous incarnation which he chose not to count and one other aborted regeneration. The name Eleventh is still used for this incarnation; the same episode depicts the prophesied "Fall of the Eleventh" which had been \ldots{}}{When did the Eleventh Doctor appear in the series the second time?}{2013}{2010} \\
        
        \synqa{} & \exampleSquadMax{Harvard's faculty includes scholars such as biologist E. O. Wilson, cognitive scientist Steven Pinker, physicists Lisa Randall and \remove{Roy Glauber}, chemists Elias Corey, Dudley R. Herschbach and George M. Whitesides, computer scientists Michael O. Rabin and \ldots{} scholar/composers Robert Levin and Bernard Rands, astrophysicist \add{Alyssa A. Goodman}, and legal scholars Alan Dershowitz and Lawrence Lessig.}{What faculty member is in a field closely related to that of Lisa Randall?}{Alyssa A. Goodman}{Roy Glauber} \\
        
        \synqa{} & \exampleSquadMax{\ldots{} and the \add{Fogg Museum of Art}, covers Western art from the Middle Ages to the present emphasizing Italian early Renaissance, British pre-Raphaelite, and 19th-century French art \ldots{} Other museums include the Carpenter Center for the Visual Arts, designed by Le Corbusier, housing the film archive, the \remove{Peabody Museum of Archaeology and Ethnology}, specializing in the cultural history and civilizations of the Western Hemisphere, and the Semitic Museum featuring artifacts from excavations in the Middle East.}{Which museum is specific to the Mediterranean cultures?}{Fogg Museum of Art}{Peabody Museum of Archaeology and Ethnology} \\
        
        \midrule
        
        \synqaext{} & \exampleSquadMax{\ldots{} In this arrangement, \add{the architect} or engineer acts as the project coordinator. His or her role is to design the works, prepare the \ldots{} There are direct contractual links between the \remove{architect's client} and the main contractor\ldots{}}{Who coordinates the project of the engineer does not?}{the architect}{architect's client} \\
        
        \synqaext{} & \exampleSquadMax{\ldots{}repoussé work and embroidery. Tibetan art from the 14th to the 19th century is represented by notable 14th- and 15th-century religious images in wood and bronze, scroll paintings and ritual objects. Art from Thailand, Burma, Cambodia, Indonesia and Sri Lanka in gold, silver, bronze, stone, terracotta and ivory represents these rich and complex cultures, the displays span the 6th to 19th centuries. Refined Hindu and Buddhist sculptures reflect the influence of India; items on show include betel-nut cutters, \add{ivory} combs and \remove{bronze} palanquin hooks.}{What material is on display with Buddhist sculptures, but not Tibetan art?}{ivory}{bronze} \\
        
        \synqaext{} & \exampleSquadMax{\ldots{}Governor Vaudreuil negotiated from Montreal a capitulation with General Amherst. Amherst granted Vaudreuil's request that any French residents who chose to remain in the colony would be given freedom to continue \ldots{} The \add{British} provided medical treatment for the sick and wounded \remove{French} soldiers\ldots{}}{What Nationality was General Amherst?}{British}{French} \\

        \bottomrule
        \end{tabular}
    \caption{
    Examples of questions that fool each of the final four models during Adversarial Human Evaluation.
    }
    \label{tab:examples_ahe}
\end{table*}
%

\end{document}